\documentclass[10pt]{article} 

\usepackage{amsmath,amsfonts,bm}

\def\eqref#1{equation~\ref{#1}}

\def\1{\bm{1}}

\DeclareMathAlphabet{\mathsfit}{\encodingdefault}{\sfdefault}{m}{sl}
\SetMathAlphabet{\mathsfit}{bold}{\encodingdefault}{\sfdefault}{bx}{n}

\usepackage{booktabs}       
\usepackage{amsfonts}       
\usepackage{nicefrac}       
\usepackage{microtype}      
\usepackage{xcolor}         

\usepackage{algorithm}
\usepackage{algpseudocode}
\usepackage{amsmath}

\newcommand{\xmark}{\ding{55}}%
\usepackage{csquotes}
\usepackage{bbm}
\usepackage{amssymb}
\usepackage{amsmath}
\usepackage{enumitem}
\usepackage{tablefootnote}
\usepackage{longtable}
\usepackage{pgfplots}
\usepackage{natbib}
\usepackage{subcaption}
\usepackage{siunitx}
\usepackage{multirow}
\usepackage{dblfloatfix}
\usepackage[usestackEOL]{stackengine}
\usepackage{wrapfig}
\usepackage{makecell}

\usepackage{dcolumn}

\newcolumntype{L}{D{.}{.}{2,3}}
\newcolumntype{M}{D{.}{.}{2,2}}

\usepgfplotslibrary{fillbetween}
\usepgfplotslibrary{colorbrewer}
    \pgfplotsset{
        cycle list/Paired-12,
        cycle multiindex* list={
            mark list*\nextlist
            Paired-12\nextlist
        },
    }
\pgfdeclarelayer{bg}
\pgfsetlayers{bg,main}

\usetikzlibrary{arrows.meta,decorations.markings}

\def\D{\mathcal{D}}

\def\X{\mathcal{X}}
\def\Y{\mathcal{Y}}

\usepackage{pifont}

\usepackage[accepted]{tmlr}

\usepackage[htt]{hyphenat}
\usepackage{hyperref}
\usepackage{url}
\usepackage{cleveref}

\title{Predicting Out-of-Domain Generalization with Neighborhood Invariance}

\author{\name Nathan Ng \email nathanng@mit.edu \\
\addr University of Toronto\\
Vector Institute\\
Massachusetts Institute of Technology
\AND
\name Neha Hulkund \email nhulkund@mit.edu \\
\addr Massachusetts Institute of Technology
\AND
\name Kyunghyun Cho \email kyunghyun.cho@nyu.edu \\
\addr New York University \\
Prescient Design, Genentech \\
CIFAR Fellow
\AND
\name Marzyeh Ghassemi \email mghassem@mit.edu \\
\addr Massachusetts Institute of Technology\\
CIFAR AI Chair\\
Vector Institute
}

\def\month{06}  
\def\year{2023} 
\def\openreview{\url{https://openreview.net/forum?id=jYkWdJzTwn}} 

\begin{document}

\maketitle

\begin{abstract}
Developing and deploying machine learning models safely depends on the ability to characterize and compare their abilities to 
generalize to new environments.
Although recent work has proposed a variety of
methods that can directly predict or theoretically bound the generalization capacity of a model, they rely on strong assumptions such as matching train/test distributions and access to model gradients.
In order to characterize generalization when these assumptions are not satisfied, we propose neighborhood invariance, a measure of a classifier's output invariance in a local transformation neighborhood.
Specifically, we sample a set of transformations and given an input test point, calculate the invariance as the largest fraction of transformed points classified into the same class.
Crucially, our measure is simple to calculate, does not depend on the test point's true label, makes no assumptions about the data distribution or model, and can be applied even in out-of-domain (OOD) settings where existing methods cannot, requiring only selecting a set of appropriate data transformations.
In experiments on robustness benchmarks in image classification, sentiment analysis, and natural language inference, we demonstrate a strong and robust correlation between our neighborhood invariance measure and actual OOD generalization
on over 4,600 models evaluated on over 100 unique train/test domain pairs.

\end{abstract}

\section{Introduction}
As deep neural networks find increasing use in safety-critical domains such as autonomous driving \citep{GUPTA2021100057} and healthcare \citep{651011},
it is important to develop methods to 
understand and compare how these models generalize to new environments.
Although empirically these models generalize in many settings \citep{hendrycks-etal-2020-pretrained, zhu2018learning, neyshabur2017exploring}
, they also exhibit numerous failure cases.
For example, models have been shown to overfit to a dataset's meta characteristics \citep{pmlr-v97-recht19a} or arbitrarily corrupted labels \citep{zhang2016understanding}, learn spurious correlations \citep{liang2022metashift}, and change their predictions even with small adversarial perturbations~\citep{goodfellow2014explaining, papernot2017practical9}.
Many methods have been proposed to mitigate these issues,  
but precisely characterizing the generalization properties of a model in diverse settings remains an open problem.


One line of work aims to theoretically bound generalization capacity~\citep{vapnik1971uniform, bartlett2003rademacher, mcallester1999pac, neyshabur2017exploring, dziugaite2017computing, pmlr-v40-Neyshabur15} or directly predict generalization~\citep{Keskar2016, pmlr-v89-liang19a, NIPS2015_eaa32c96, schiff2021predicting, jiang2018predicting},
and are useful in reasoning about a model beyond its performance on a specific known test set.
However, these methods work only when train and test distributions are the same, and 
often rely on a strong set of assumptions such as access to labelled test data \citep{schiff2021predicting}, model weights \citep{pmlr-v40-Neyshabur15,bartlett2017spectrally,neyshabhur2017pac}, model gradients \citep{jiang2018predicting}, and training data \citep{Keskar2016}.
More recent work aims to estimate the generalization of a trained model on unlabelled test data directly~\citep{deng2020labels, 2021arXiv210613799J, Deng:ICML2021, 2022arXiv220104234G}. 
However, these metrics are typically calculated based on the output logits of a model on individual examples, which can become poorly calibrated in out-of-domain (OOD) settings \citep{morteza2022provable}.
In real world settings, we require a robust measure of generalization that can be applied across a wide range of test distributions and where we are often given access only to a black box model.

\begin{figure}
    \centering
    \includegraphics[scale=0.19]{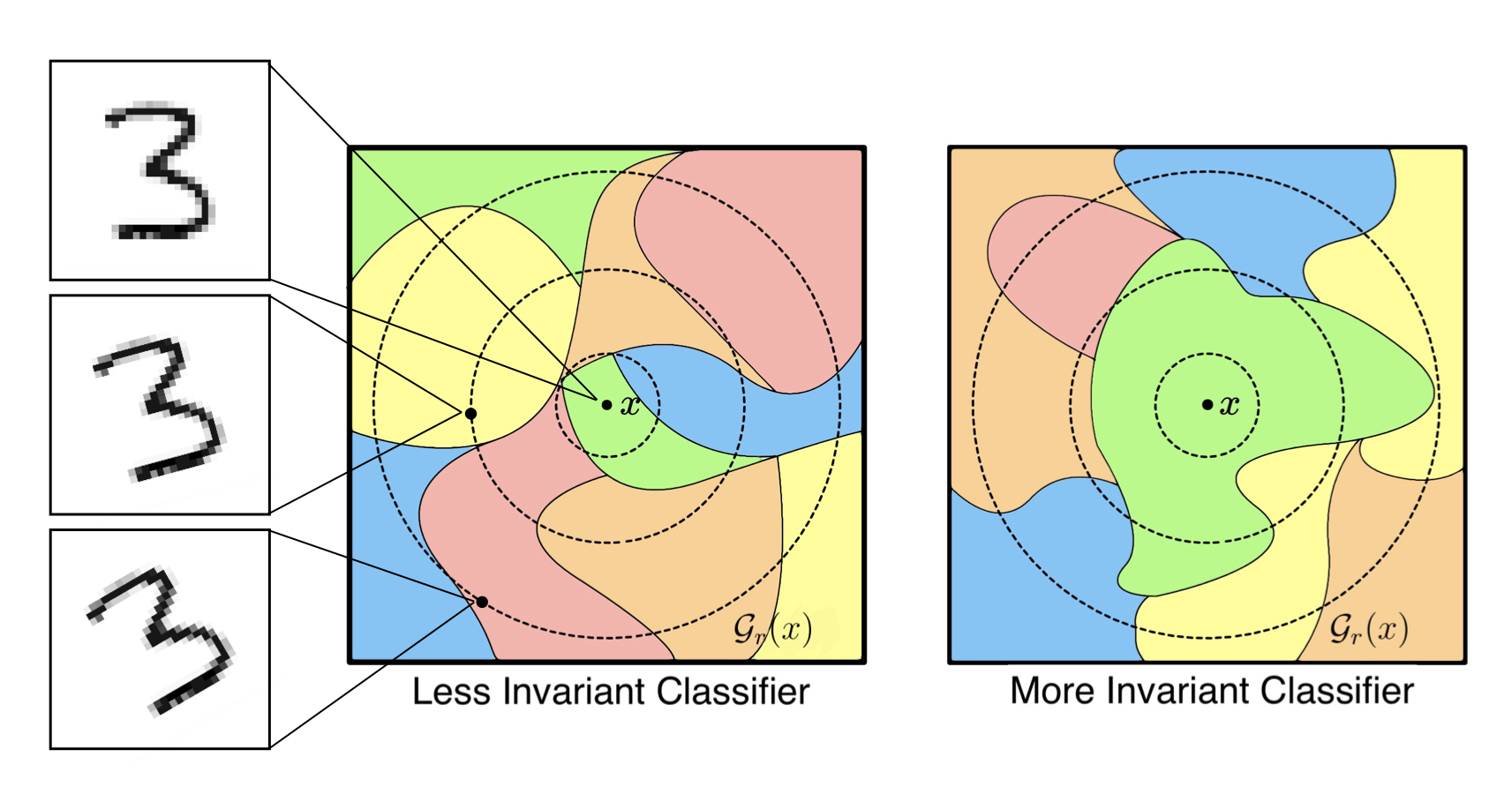}
    \caption{
    The transformation neighborhood $\mathcal{G}_r(x)$ around $x$ contains the set of points reachable from a transformation in $\mathcal{G}_r$ (pictured here as a rotation of $r$ degrees). As $r$ increases, the transformation neighborhood is partitioned into a distinct set of decision regions.
    More invariant classifiers (right) classify most points in the neighborhood into the same class and should generalize better compared to less invariant classifiers (left).
    Our measure of invariance is independent of what specific region $x$ lies in.
    }
    \label{fig:neighborhood}
\end{figure}

In this paper we propose \textit{neighborhood invariance}, a complexity measure that correlates well with generalization and that only assumes access to a set of suitable data transformations.
Given a test data point, we define the transformation neighborhood as the set of points that can be generated from a set of transformations with a given maximum magnitude.
A classifier's neighborhood invariance is then the proportion of points that are classified into the most commonly predicted class in this neighborhood.
Intuitively, a classifier that is more invariant in this neighborhood should have be able to represent examples with lower dimensionality and thus lower complexity, leading to stronger generalization.
Different from other similar methods~\citep{aithal2021robustness}, we 
define invariance with respect to the neighborhood itself rather than relative to the prediction at the test point and do not require manually tuning weights, meaning our measure can be applied even when test distributions vary.
In addition, since our measure makes so few assumptions it is applicable in a wide range of experimental settings and can be used to compare the generalization properties of multiple models even when labeled data is unavailable.



We investigate the correlation of a model's neighborhood invariance with its capacity to generalize, focusing on experimental settings with OOD dataset shifts \citep{taori2020measuring} where test data is sampled from a distribution different from the training distribution. 
We select common OOD benchmark datasets in image classification \citep{Krizhevsky2009learning, lu2020cifar10.2, recht2018cifar10.1, deng2012mnist, darlow2018cinic10, Netzer2011ReadingDI, InvariantRiskMinimization, taori2020measuring}, sentiment analysis \citep{jianmo2019justifying}, and natural language inference \citep{williams2018broad}, which totals over 100 pairs of training/test domains.
We consider a large pool of over 4,600 models trained on these datasets with varying architectures and generalization properties, and sample sets of transformations commonly used for data augmentation \citep{ng2020ssmba, cubuk2020randaugment, wei-zou-2019-eda, xie2019unsupervised}. 
Across a wide set of correlation metrics, we find that neighborhood invariance measures outperform or match baselines in almost all experimental settings.
\section{Related Work}

\textbf{Characterizing Model Invariance}\quad 
Ensuring various kinds of model invariance is a well studied aspect of learning generalizable models and has been analyzed extensively from a causality perspective \citep{bühlmann2018invariance, peters2015causal, 10.2307/1905714}. 
At the largest scale, models trained on a wide support of training data and domains have demonstrated robust zero-shot and few-shot abilities \citep{radford2021clip, brown2020gpt3,wortsman2022robust}.
At a smaller scale, 
models that are invariant across data domains or interventions \citep{InvariantRiskMinimization, gulrajani2020insearch, bühlmann2018invariance} are able to learn representations that do not depend on spurious correlations. 
Finally, at the smallest scale, local invariance to data augmentations \citep{cubuk2020randaugment},
local changes \citep{rifai2011manifold}, augmentation graphs \citep{haochen2021provable}, similar neighbors \citep{luo2018smooth}, or interpolation between points \citep{chaudhuri2019manifold, zhang2018mixup} have demonstrated improvements in model generalization.
In our paper, we consider model invariance at this local scale.

Recent work has shown that models that are invariant to local transformations factorize the input space into a base space and the set of transformations \citep{8024476, sannai2021improved}, 
effectively reducing the input dimensionality and thus model complexity \citep{ANSELMI2016112, anselmi2015invariance}.
Measuring this decrease in complexity can be performed by analyzing the sample cover \citep{zhu2021understanding}.
A similar line of work derives estimation error bounds based on the intrinsic dimensionality of deep ReLU networks in H{\"o}lder \citep{schmidt2019deep, ryumei2020adaptive,chen2019nonparametric}, Besov, mixed smooth Besov \citep{suzuki2018adaptivity}, and anisotropic Besov \citep{suzuki2021deep} function spaces.

Most similar to our work, \citet{aithal2021robustness} measures a model's robustness to perturbations as a proxy for generalization. 
Our method generalizes theirs and differs in a few key ways.
We calculate our measure on the test set relative to a transformation neighborhood and can thus adapt to any specific domain for which we measure complexity and predict generalization. In contrast, \citet{aithal2021robustness} calculate their measure on the training set, use the models' prediction as a ground truth, and require manually tuning the weights of augmentations, meaning it is relatively brittle and can only be applied to in-domain data.
In addition we analyze the correlation of our neighborhood invariance measure on a wide range of OOD benchmarks on image classification, sentiment analysis, and natural language inference, while \citet{aithal2021robustness} consider only image classification tasks with matching train/test distributions.


\textbf{Measures of Complexity and Predicting Generalization} \quad
Traditional methods of analyzing the generalization bounds of neural networks use theoretical measures of complexity.
VC dimension \citep{vapnik1971uniform} and Rademacher complexity \citep{bartlett2003rademacher} can be used to bound the generalization of particular function classes, although they are often vacuous at the scale of deep neural networks \citep{dziugaite2017computing}.
The PAC-Bayes framework \citep{mcallester1999pac, neyshabur2017exploring, dziugaite2017computing, pmlr-v130-garg21a}
can be used to build tighter generalization bounds by considering the ``sharpness'' of the local minima.
Norm-based measures \citep{pmlr-v40-Neyshabur15,bartlett2017spectrally,neyshabhur2017pac} bound generalization by considering different norms of the weights of learned networks.
More recent analyses have focused on empirically motivated measures that do not provide theoretical bounds.
These include 
the sharpness of minima in parameter space \cite{Keskar2016}, 
Fisher-Rao norm \cite{pmlr-v89-liang19a}, distance from initialization \citep{Nagarajan_Kolter_2019}, path norm \citep{NIPS2015_eaa32c96}, layer margin distributions \citep{jiang2018predicting}, and perturbation response curves \cite{schiff2021predicting}.

However, these measures are only applicable when train and test distributions match.
Although some generalization bounds have been derived for these OOD settings \citep{pmlr-v130-garg21a, bendavid2007analysis, zhang2019bridging}, they rely on access to the test data distribution.
In addition, many testbeds examine only synthetic shifts, whereas natural shifts such as WILDS \citep{wilds2021} are much more difficult.
In real world settings where test distributions are often unknown, a separate line of work aims to directly predict generalization from unlabelled test data.
These methods either predict the correctness on individual examples \citep{deng2020labels, 2021arXiv210613799J, Deng:ICML2021}, directly estimate the total error \citep{2022arXiv220104234G, guillory2021predicting, chenmandoline2021, chuang2020estimating, ramarkishna2021empirical}, or learn linear models relating ID and OOD accuracy \citep{miller2021accuracy} or agreement \citep{baek2022agreementontheline}.


\section{Neighborhood Invariance Measure}
\label{sec:smoothness}

In this section we introduce our neighborhood invariance measure.
We start by defining the transformation neighborhood of a point, then motivate our formulation of invariance in this neighborhood, and finally show how to estimate it in practice.

\begin{figure}
\centering
\includegraphics[scale=0.3]{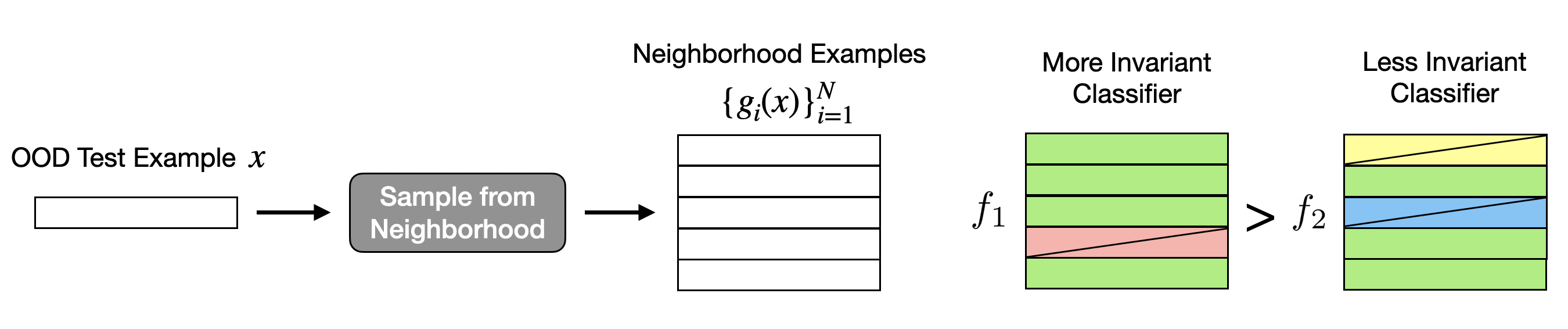}
\caption{To estimate neighborhood invariance we sample a set of transformations $\{g_i\}_{i=1}^N \sim \mathcal{G}_r$ and generate a
set of nearby examples $\{g_i(x)\}_{i=1}^N$ for every test example $x$. This set of examples is then evaluated using each classifier. We expect classifiers with more points classified in the most common class to generalize better to the given test set.}
\label{fig:measure_graph}
\end{figure}

\subsection{Motivation}
Consider a classification task from an input space $\X$ 
to an output space $\Y$ with $k$ classes.
We are given a model $f: \X \rightarrow \Y$ trained on 
an in-domain training dataset $\D_i = \{(x_i^{1}, y_i^{1}), \ldots, (x_i^{n}, y_i^{n})\}$ sampled from a distribution $P_i(\X, \Y)$, and an out-of-domain test dataset $\D_o = \{(x_o^{1}, y_o^{1}), \ldots, (x_o^{m}, y_o^{m})\}$ sampled from a distribution $P_o(\X, \Y)$.
We assume further that domains are covariate shifted such that $P(\Y|\X)$ does not change between domains 

We consider a set of data transformations $\mathcal{G}_r = \{g: \X \rightarrow \X \mid m(g) < r\}$ where $g$ is a particular data transformation with an associated measure of the magnitude of the transformation $m(g)$. 
For example, a set of image rotation transformations with a maximum angle of 30 degrees might be denoted $\mathcal{G}_{30}$ where $m(g_i) = \alpha$ is the angle of rotation for a specific $g_i$.
For a given test point $x \in \D_o$, we define the transformation neighborhood $\mathcal{G}_r(x) = \{g(x) \in \X \mid g \in \mathcal{G}_r\}$ as the set of outputs after applying all transformations in $\mathcal{G}_r$. 
Defining the neighborhood this way allows us to consider a wide range of nearby points in a controllable way without 
needing access to the underlying data distribution.
As shown in Figure \ref{fig:neighborhood}, for a given $r$, we can define a neighborhood decision distribution as 
\begin{align}
    p_j(x) = \frac{|\{f(x') = j \mid x' \in \mathcal{G}_r(x)\} |}{|\mathcal{G}_r(x)|}.
    \label{eq:smoothness}
\end{align}
We then define our \textbf{neighborhood invariance measure} as 
\begin{align}
    \mu(f, x) = \max_{j \in \Y} p_j(x). 
    \label{eq:dominant_label}
\end{align}

We assume that data transformations are selected such that the label for the transformed point $g(x)$ is still well defined. 
For example, flipping MNIST digits horizontally would cause most examples to have an undefined label.
If the label is undefined, then $f(g(x))$ should produce close to random outputs and thus constant invariance values of $\nicefrac{1}{k}$ regardless of the generalization properties of $f$, causing our measure to fail. 
We empirically measure this phenomenon across a wide range of transformations in Section \ref{sec:ablations}.

Intuitively, a classifier that is more invariant in the neighborhood of $x$ should be able to represent it with a lower input dimensionality and thus be less complex, leading to stronger generalization capabilities.
In contrast, a less invariant classifier will need a higher input dimensionality to represent the neighborhood of $x$ and thus be more complex, leading to weaker generalization.
Crucially, since our invariance is measured with respect to the neighborhood around the test point rather than the test point itself, it does not rely on the ground truth label. 
In addition, it makes no assumptions about the model or the distribution from which test data was sampled, making it applicable in many settings where existing complexity measures cannot be calculated, including common OOD robustness settings.

\subsection{Estimating Neighborhood Invariance}
Calculating neighborhood invariance exactly is typically intractable since evaluating all possible transformations is impossible. 
Instead, we perform Monte Carlo estimation by sampling a set of $N$ transformations $\{g_i\}_{i=1}^N$ from $\mathcal{G}_r$ (including the identity transformation $I(x) = x$) and calculating
\begin{align}
    \mu(f, x) =  \frac{1}{N} \sum_{i=1}^N \mathbbm{1}\left(f(g_i(x)) = \hat{y}(f, x) \right),
    \label{eq:smoothness_approx}
\end{align}
where
\begin{align}
\hat{y}(f,x) = \arg\max_{j \in \Y} \sum_{i=1}^N \mathbbm{1}\left( f(g_i(x)) = j \right)
\label{eq:dominant_label_approx}
\end{align}
is the most commonly output label.
The average neighborhood invariance across the entire dataset $\D_o$ is then $\frac{1}{m}\sum_{j=1}^m \mu(f, x_o^j)$.

\section{Experimental Setup}

Empirically evaluating the quality of a complexity measure is difficult and requires careful experimental design.
Typically, evaluation is done by generating a large pool of models with sufficiently varied generalization properties, but if we generate these models by varying only a few hyperparameters, our observed correlation may be an artifact of these factors affecting both generalization and our measure.
To this end, we follow a similar experimental setup to \citet{jiang2019fantastic}.

\subsection{Data}
\label{subsec:data}

For our experiments we focus on three tasks: large and small scale image classification, 
sentiment analysis on single sentences and natural language inference on sentence pairs. For each task we construct a set of datasets sampled from different data domains.

\textbf{Image Classification} \quad 
For image classification we begin by considering 7 datasets domain shifted from ImageNet \citep{deng2009imagenet,russakovsky2015ilsvrc} 
These include ImageNetV2 \citep{pmlr-v97-recht19a} and Imagenet-Sketch \citep{wang2019imagenetsketch} with the same output classes, as well as ObjectNet \citep{barbu2019objectnet}, ImageNetVid, YTBB anchors \citep{gu2019using, pmlr-v97-recht19a}, ImageNet-A \citep{hendrycks2021nae},  and ImageNet-R \citep{hendrycks2021many} with a smaller subset of output classes.

In addition to ImageNet datasets we construct two sets of smaller scale datasets. The first we call \textbf{CI10} and consists of CIFAR10 \citep{Krizhevsky2009learning}, CINIC10 \citep{darlow2018cinic10}, CIFAR10.1\citep{recht2018cifar10.1}, and CIFAR10.2\citep{lu2020cifar10.2}.
The second we call \textbf{Numbers} and consists of SVHN \citep{Netzer2011ReadingDI}, MNIST \citep{deng2012mnist}, and Colored MNIST \citep{InvariantRiskMinimization}. Domains in each set share the same set of output classes.

\textbf{Sentiment Analysis (SA)} \quad We use the datasets subsampled from the Amazon reviews dataset \citep{jianmo2019justifying} which contains product reviews from Amazon. 
Following \citet{hendrycks2020pretrained} and \citet{ng2020ssmba}, we split the dataset into 10 different domains based on review category. 
For all domains and datasets, models are trained to predict a review's star rating from 1 to 5.

\textbf{Natural Language Inference (NLI)} \quad We use the MNLI \citep{williams2018broad} dataset, a corpus of NLI data from 10 distinct genres of written and spoken English.
We train on the 5 genres with training data and evaluate on all 10 genres.
Models are given two sentences, a premise and hypothesis, and predict whether the hypothesis is entailed by, is neutral to, or contradicts the premise.

\subsection{Model and Hyperparameter Space}
\label{subsec:models}
\begin{table*}[t]
\small
\centering
\resizebox{\columnwidth}{!}{%
\begin{tabular}{llccccccccccccc}
\toprule
\textbf{Model} & \textbf{Dataset} & \thead{Training \\ Domain} & \thead{Batch \\Size} & \thead{Depth} & \thead{Width} & \thead{Dropout} & \thead{Weight \\ Decay} & \thead{Label \\ Noise} & \thead{Learning \\ Rate} & \thead{Batch\\Norm} & \thead{Seed} & \thead{Data\\Aug} & \thead{\# Converged} & \thead{\# Evaluations}\\
\midrule
\multirow{2}{*}{CNN} & Amazon & 10 & 3 & 3 & 3 & 3 & 3 & --- & --- & --- & --- & --- & 2,418 & 24,180\\
& MNLI & 5 & 3 & 3 & 3 & 2 & --- & 3 & --- & --- & --- & --- & 796 & 39,800\\
\multirow{2}{*}{RoBERTa} & Amazon & 10 & 3 & --- & --- & 2 & 3 & 3 & --- & --- & --- & --- & 332 & 33,200 \\
& MNLI & 5 & 3 & --- & --- & 2 & 3 & 3 & --- & --- & --- & --- & 213 & 10,650 \\
\midrule
Various & ImageNet & --- & --- & --- & --- & --- & --- & --- & --- & --- & --- & --- & 401 & 2,406\\
\multirow{2}{*}{NiN} & SVHN & --- & 3 & 3 & --- & 3 & 2 & --- & --- & --- & --- & --- & 54 & 162 \\
& CIFAR10 & --- & 2 & 2 & 2 & 2 & 2 & --- & --- & --- & --- & --- &  32 & 128\\
VGG & CIFAR10 & --- & 3 & 3 & --- & 3 & 2 & --- & --- & --- & --- & --- & 54 & 216\\
ResNet & CIFAR10 & --- & --- & --- & 3 & --- & 3 & --- & 2 & 2 & 3 & 2 & 216 & 864\\ 
CNN & CINIC10 & --- & 2 & 2 & 4 & --- & 2 & --- & 2 & 2 & --- & --- &  128 & 512 \\
\cmidrule(lr){1-13}\cmidrule(lr){14-14}\cmidrule(lr){15-15}
& & & & & & & & & & & & & 4,644 & 112,118\\
\bottomrule
\end{tabular}
}
\caption{Number of possible hyperparameter values for each architecture and task. Fields denoted with a --- indicate that this hyperparameter is fixed or not applicable. We also list the total number of models converged and evaluations run in each model pool. In total we consider 4,644 models and 112,118 evaluations. }
\label{tab:hyperparams}
\end{table*}

For large scale image classification on ImageNet, we use pretrained models from the ImageNet Testbed \citep{taori2020measuring} which covers a wide range of architectures including ResNext \citep{Xie2016resnext}, EfficientNet \citep{tan2019efficientnet}, BiT \citep{beyer2021bit}, Vision Transformers \citep{dosovitskiy2020vit}, CLIP \citep{radford2021clip}, and many more models.
We provide a full list of models evaluated in Appendix \ref{app:app_arch}.
For smaller scale image classification tasks, we use models trained for the tasks 1, 2, 4, 5, and 9 from the Predicting Generalization in Deep Learning competition (PGDL) \citep{jiang2020predicting} as well as models from \citet{jiang2018predicting}, which covers Network in Network (NiN) \citep{lin2013network}, VGG \citep{simonyan2015very}, ResNet \citep{he2015resnet}, and CNN models trained on CIFAR10, CINIC10, and SVHN.
On natural language tasks we consider CNN \citep{kim2014convolutional,mou-etal-2016-natural} and RoBERTa \citep{liu2019roberta} based models.
On natural language models we apply label noise by randomly replacing a fraction of training labels with uniform samples from the label space.
We argue that label noise is not an artificial training setting as stated in \cite{jiang2019fantastic} but rather a method of entropy regularization \citep{pereyra2017regularizing, 2016arXiv160500055X} which prevents models from becoming overconfident.

In order to control for the varying convergence rates and learning capacities of our different models, we follow \cite{jiang2019fantastic} and early stop the training of models when they reach a given training cross entropy loss (usually around 99\% training accuracy), or if they reach the max number of training epochs.
We discard all models which do not converge within this time.
The total number of models trained and converged in each pool as well as details on hyperparameter variations for each task and model provided in Table \ref{tab:hyperparams}.
We include further details on model training, the hyperparameter space, and specific choices in hyperparameters in Appendix \ref{app:model_train}, \ref{app:app_arch}, and \ref{app:model_hyperparams}.

\subsection{Evaluation Metrics}
\label{sec:metrics}
Given a set of domains defined by distributions $\{P_1, P_2, \ldots P_n\}$ and a set of datasets $\{\D_i \sim P_i\}_{i=1}^n$ sampled from these domains, we train a set of models $F_i = \{f_i^1, f_i^2, \ldots, f_i^m\}$ on each dataset $\D_i$.
We evaluate all models $f_i^k \in F_i$ on all OOD test datasets $\D_o : o \neq i$, generating a set of invariance and generalization values $(\mu_{io}^k, g_{io}^k)$.
We define generalization as the top-1 accuracy of $f_i^k$ on $\D_o$.

We evaluate our measure first by 
predicting the generalization of a given model to an OOD test set.
Specifically, we select an OOD test set $\D_o$ and an in-domain training set $\D_i : i \neq o$ and predict the OOD generalization $g_{io}^k$ of a model $f_i^k$ trained on $\D_i$ and evaluated on $\D_o$ from its invaraince value $\mu_{io}^k$. 
To generate these predictions we use a linear model $\hat{g} = a\mu + b$ with parameters $a, b \in \mathcal{R}$.
To estimate our parameters $a$ and $b$, we select a pool of models $\{f_j^k \in F_j : j \neq i,o \}$ that are trained on \emph{all remaining datasets}.
Each model $f_j^k$ in this pool is evaluated on the OOD dataset $\D_o$ to give us a set of pairs $\{(\mu_{jo}^k, g_{jo}^k)\}$. 
We then find $a, b$ by minimizing the mean squared error $(a^*, b^*) = \arg \min_{a, b} \sum_{j,k} (a\mu_{jo}^k + b - g_{jo}^k)^2$ on all models in the pool.

We use the learned parameters to make generalization predictions $\hat{g}_{io}^k = a\mu_{io}^k + b$ for every model $f_i^k \in F_i$ on $\D_o$
and measure the coefficient of determination $\mathbf{R^2}$ \citep{glantz1990primer}.
We also measure the residuals of our linear model by calculating the mean absolute error (\textbf{MAE}) between our predictions and the actual generalization.
For every pair of training domain $i$ and OOD test domain $o$, we evaluate $R^2$ and MAE then average each metric across all pairs. We report MAE values as percentage points.


We also consider the rank correlation between neighborhood invariance and actual generalization.
Specifically, for a pair of models $f_i, f_j$ with measure and generalization pairs $(\mu_i, g_i)$ and $(\mu_j, g_j)$, we want $g_i > g_j$ if $\mu_i > \mu_j$.
We use Kendall's rank $\tau$ coefficient \citep{kendall1938new} to measure how consistent these sets of rankings are.
We measure four different $\tau$ values:

\textbf{ID $\tau$} \quad This metric evaluates the correlation of our measure with in-domain generalization.
We select a training dataset $\D_i$ and consider pairs $\{(\mu_{ii}^k, g_{ii}^k)\}$ generated from the set of models $F_i$ trained on $\D_i$. 
$\tau$ values are averaged across all training domains.

\textbf{Macro $\tau$} \quad This metric evaluates the correlation of our measure individually on each training/OOD test domain pair. 
We select a training dataset $\D_i$ and a OOD test dataset $\D_o$ and consider pairs $\{(\mu_{io}^k, g_{io}^k)\}$ generated from the set of models $F_i$ trained on $\D_i$.
$\tau$ values are averaged across all pairs of training and OOD test domains. 

\textbf{Micro $\tau$} \quad This metric evaluates the correlation of our measure on a given OOD test domain across models trained on all other domains.
We select a single OOD test domain $\D_o$ and consider pairs $\{(\mu_{io}^k, g_{io}^k)\}$ generated from the set of models $\{f_i^k \in F_i : i \neq o\}$ trained on all other datasets $\{D_i : i \neq o \}$.
$\tau$ values are averaged across all test domains. 
We use this metric only when different models are trained on different training sets.


\textbf{Arch $\tau$} \quad This metric evaluates the correlation of our measure on models trained with different architectures.
Arch $\tau$ is calculated similar to Micro $\tau$, except $F_i$ now includes models from all architectures.
$\tau$ values are averaged across all test domains.

\subsection{Data Transformations}
Defining the transformation neighborhood requires defining a set of data transformations with associated magnitudes.
For image classification, we consider four transformations: RandAugment \citep{cubuk2020randaugment} which randomly combines various transformations, random translation in the X- and Y-axes, random patch erasing \citep{zhong2020random} which removes randomly sized patches from the image, and horizontal flips and crops.
We call neighborhood invariance measures based on these transformations \textbf{NI-RandAug}, \textbf{NI-Translate}, \textbf{NI-Erase}, and \textbf{NI-FC} respectively.
For natural language tasks, we also consider four transformations: SSMBA \citep{ng2020ssmba} which generates examples in a manifold neighborhood using a denoising autoencoder, EDA \citep{wei-zou-2019-eda} which applies random word level operations, backtranslation (BT) \citep{sennrich2016improving, xie2019unsupervised} which translates back and forth from a pivot language, and a transformation that randomly replaces a percentage of tokens. 
We call neighborhood invariance measures based on these transformations \textbf{NI-SSMBA}, \textbf{NI-EDA}, \textbf{NI-BT}, and \textbf{NI-RandRep} respectively.
For all experiments we sample $n=10$ transformations in addition to the identity tranfsormation, although ablations in section \ref{sec:ablations} show that our method is relatively robust to the specific number of transformations sampled.
We provide further details on specific transformation magnitude values and implementations for all methods in Appendix \ref{sec:app_neighborhood_hyperparams}.


\subsection{Baselines}
\label{sec:baselines}
Since our experimental setting makes so few assumptions, there are very few complexity measures that we can compare against.
This includes \cite{aithal2021robustness}, which requires matching train/test distributions.
We thus consider complexity measures that require only model weights, specifically the \textbf{Spectral} \citep{yoshida2017spectral, neyshabhur2017pac} and \textbf{Frobenius} \citep{pmlr-v40-Neyshabur15} norms.
However, in our experiments we find close to 0 or negative correlation for these measures, so we do not report their performance.
We also compare our method against output based methods that directly predict OOD generalization.
We use \textbf{ATC-MC} and \textbf{ATC-NE} \cite{2022arXiv220104234G} as our two baselines, which calculate a threshold on in-domain validation data based on max confidence and negative entropy scores respectively.
To calculate metrics on these methods we treat the generated accuracy predictions as a score.
To calculate ID $\tau$ values we select a threshold value based on validation data then calculate predicted accuracy values on test data from the same domain.
For ImageNet domain shift datasets where the output classes are a subset of the original 1,000 ImageNet classes, we do not recompute a subclassed ATC threshold as we do not assume prior knowledge of the OOD output classes.

\section{Results}
\label{sec:results}
We now present the results of our experiments evaluating the quality of our neighborhood invariance measure. 
We begin by analyzing the effect of dataset distance on the correlation of neighborhood invariance with generalization in a toy setting.
Our main set of results evaluate neighborhood invariance on OOD benchmarks in image classification, sentiment analysis, and natural language inference. 
Finally, we examine the correlation of our measure in extreme OOD settings and analyze the factors that affect the quality of our neighborhood invariance estimates.

\subsection{Dataset Distance: Toy Analysis}
\label{sec:toy}
In general, as with any complexity measure or generalization predictor, we expect neighborhood invariance to perform more poorly as we move farther from the training domain.
In the worst case, if a classifier becomes a degenerate constant classifier in a far enough OOD domain, then neighborhood invariance reaches a constant maximum value of 1 while generalization becomes random.
In order for our measure to work well, we assume that test domains are sufficiently close to training domains so that model predictions are non-constant.
In this section we present an analysis of the effect of dataset distance on the quality of our measure in a toy setting.

We consider a binary classification task of points inside and outside a unit hypersphere in $\X = \mathcal{R}^{n}$, as shown in Figure \ref{fig:measure_graph}. 
We define different data domains as univariate gaussian distributions $P(\X)$, centered at a point $\mu$ on the hypersphere. 
The distance between two datasets $\D_1\sim P_1(\X)$ and $\D_2 \sim P_2(\X)$ can then be measured as the distance along the hypersphere between $\mu_1$ and $\mu_2$ in radians.
We generate a training dataset $\D_0$ by sampling points around the north pole of the hypersphere, then generate out-of-domain datasets $\D_j$ at varying distances from $\D_0$.
Given a model trained on $\D_0$, we can calculate its generalization to $\D_j$, as well as its neighborhood invariance. 

\begin{figure}[t]
     \centering
     \begin{subfigure}[t]{0.47\textwidth}
\centering
\begin{tikzpicture}
  \draw (0,0) circle (2cm);
  \draw (0,0) circle (2cm);
  \draw (-2,0) arc (180:360:2 and 0.4);
  \draw[dashed] (2,0) arc (0:180:2 and 0.4);

  \shade[ball color = gray!40, opacity = 0.4] (0,2cm) circle (0.4cm);
  \fill[fill=black] (0,2cm) circle (1.5pt);
    \node[label=above:{$\D_0$}] at (0,2) {};

  \shade[ball color = gray!40, opacity = 0.4] (-0.6,1.62cm) circle (0.4cm);
  \fill[fill=black] (-0.6,1.62) circle (1.5pt);
  \draw (-1,1.73) arc (180:360:1 and 0.15);
  \draw[dashed] (1,1.73) arc (0:180:1 and 0.15);
      \node[label=left:{$\D_1$}] at (-1,1.73) {};

  \shade[ball color = gray!40, opacity = 0.4] (1.1,0.78) circle (0.4cm);
  \fill[fill=black] (1.1,0.78) circle (1.5pt);
  \draw (-1.74,1) arc (180:360:1.74 and 0.3);
  \draw[dashed] (1.74,1) arc (0:180:1.74 and 0.3);
      \node[label=left:{$\D_2$}] at (-1.73,1) {};

  \shade[ball color = gray!40, opacity = 0.4] (0.2,-0.38) circle (0.4cm);
  \fill[fill=black] (0.2,-0.38) circle (1.5pt);
    \node[label=left:{$\D_3$}] at (-2,0) {};

  \draw (-1,-1.73) arc (180:360:1 and 0.15);
  \draw[dashed] (1,-1.73) arc (0:180:1 and 0.1);
    \node[label=left:{$\D_5$}] at (-1,-1.73) {};

  \draw (-1.74,-1) arc (180:360:1.74 and 0.3);
  \draw[dashed] (1.74,-1) arc (0:180:1.74 and 0.3);
    \shade[ball color = gray!40, opacity = 0.4] (-0.5,-1.29) circle (0.4cm);
  \fill[fill=black] (-0.5,-1.29) circle (1.5pt);
  \node[label=left:{$\D_4$}] at (-1.74,-1) {};

  \shade[ball color = gray!40, opacity = 0.4] (0.1,-1.87) circle (0.4cm);
  \fill[fill=black] (0.1,-1.87) circle (1.5pt);

\end{tikzpicture}
\caption{We generate data from univariate gaussians whose means lie on a hypersphere. We train models on a dataset $\D_0$ to classify points as inside or outside the hypersphere then test them on out of domain datasets $\D_j$ that lie at various distances from $\D_0$ as measured by radiance distance along the hypersphere.}
\label{fig:toy_gen_data}
\end{subfigure}
\hfill
\begin{subfigure}[t]{0.47\textwidth}
    \centering
\begin{tikzpicture}[scale=0.85, trim axis left]
\begin{axis}[
    xlabel=Dataset Distance,
    ylabel=Kendall $\tau$,
    xtick pos=left,
    ytick pos=left,
    ymajorgrids=true,
    xmin=0,xmax=40,
    ymin=-1,ymax=1,
    legend pos=north east,
    height=6cm, width=6.5cm,
    xtick={0,10,20,30,40},
    xticklabels={0,$\nicefrac{\pi}{4}$,$\nicefrac{\pi}{2}$,$\nicefrac{3\pi}{4}$,$\pi$}
]


\addplot+[blue, mark options={blue}, mark size=1pt] table [x=x, y=y, col sep=comma] {
x, y
1,0.6525258532135789
2,0.5830753611787677
3,0.45543136953204966
4,0.3744321562149508
5,0.24657020219981124
6,0.16553698718211124
7,0.18565558926256362
8,0.19396307244029426
9,0.1364864439973082
10,0.03217010429856841
11,-0.012303735318492073
12,-0.03894219069435641
13,-0.01939445522468155
14,-0.08266358111144798
15,-0.03717861509776703
16,-0.13671343303864053
17,-0.1498124792646949
18,-0.2525740586689544
19,-0.28645503604304184
20,-0.32779990005381376
21,-0.29375629068200576
22,-0.3841272995754502
23,-0.36904584564920917
24,-0.37700072347301167
25,-0.4797002655215846
26,-0.46268958172504415
27,-0.5603306585380293
28,-0.5969839914629598
29,-0.5709959115485879
30,-0.6489042029453421
31,-0.6344588008283127
32,-0.6987639382996471
33,-0.7376043095802051
34,-0.7287478253359547
35,-0.739928694083101
36,-0.7558163977982721
37,-0.7318418788241242
38,-0.7576083242534059
39,-0.7664748771610461
40,-0.7269193211206649
};
\addlegendentry{NI}

\addplot+[red, mark=*, mark options={red}, mark size=1pt] table [x=x, y=y, col sep=comma] {
x,y
1,0.6176152287658037
2,0.5368563095778853
3,0.4261757336977026
4,0.31839032830471503
5,0.20612708238947558
6,0.09219225239811502
7,0.0949683718931356
8,0.1254607355949024
9,0.09078205362121733
10,0.02296657018928826
11,-0.03256325454538281
12,-0.04471917424987253
13,-0.032525174088703236
14,-0.06631935206519064
15,-0.0321232932609765
16,-0.13783610157635465
17,-0.14547467271276174
18,-0.22295090869069356
19,-0.27228480132059085
20,-0.31673951851741206
21,-0.27785419674412337
22,-0.3833009737092
23,-0.349260825079429
24,-0.35041238425906446
25,-0.44590152768199937
26,-0.42989184610688075
27,-0.5313357981495975
28,-0.5518072452257601
29,-0.5352931454008542
30,-0.6094983229875618
31,-0.602507277250487
32,-0.6539660615227365
33,-0.672656153785861
34,-0.6746579716954092
35,-0.6759466666310082
36,-0.6779893923608891
37,-0.6804479454039745
38,-0.6900451202848992
39,-0.7030530208638897
40,-0.6769555883215527
};
\addlegendentry{ATC};

\end{axis}
\end{tikzpicture}
\caption{The correlation between neighborhood invariance and OOD generalization remains high near the training domain but quickly decreases as dataset distance incerases, becoming negatively correlated for datasets further than $\nicefrac{\pi}{4}$ away on the hypersphere. We observe almost identical results for baseline ATC methods.}
\label{fig:toy_res}
\end{subfigure}
\caption{Toy analysis of the effect of dataset distance on the correlation between neighborhood invariance and OOD generalization.}
\label{fig:toy_res_full}
\end{figure}
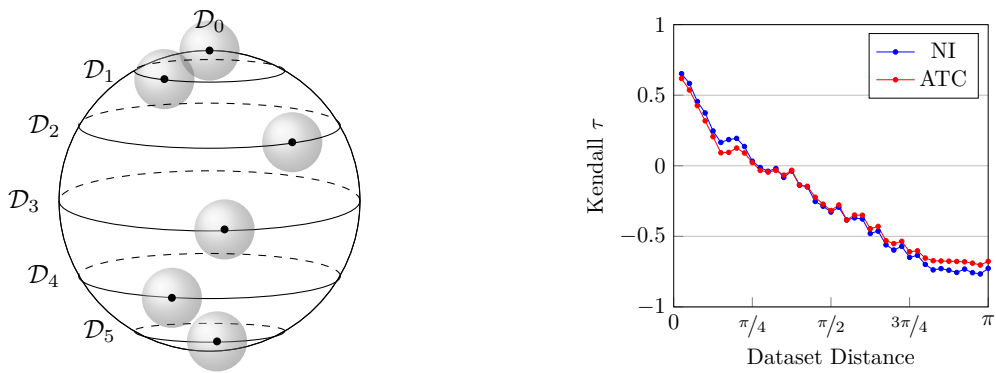

In our experiments, we consider a 16 dimensional hypersphere and sample 1000 points per data distribution for each dataset. 
The univariate gaussian distributions that we sample data points from have fixed variance 0.005, ensuring models cannot generalize fully across the entire hypersphere.
We train 200 single hidden-layer MLPs with hidden dimension of 16 on the training dataset $\D_0$, each with a random level of label noise between 0-30\% to ensure a wide range of generalization properties.
We consider 40 different dataset distances, equally spaced along the hypersphere between opposite poles. 
For a given dataset distance, we select 5 points at random from the corresponding circumference and generate 5 datasets from univariate gaussians centered at these points.
To measure neighborhood invariance for a given $x$, we sample 10 transformations from the set of transformations defined as a perturbation along the hypersphere of radius $||x||$ with a maximum distance $m(g) = ||g(x) - x|| \leq 0.01$.
For each dataset we measure the neighborhood invariance and generalization for each of the 200 trained models and calculate the Kendall $\tau$ correlation between them.
For a given dataset distance, the $\tau$ values are then averaged across all datasets.
Results are presented in Figure \ref{fig:toy_res}.

For datasets closest to the training dataset, the correlation between generalization and neighborhood invariance is high. However, as dataset distance increases, correlation decreases.
At a distance of around $\pi/4$, correlation becomes nearly $0$, and continues decreasing until the two values are negatively correlated on data sampled from the opposite side of the hypersphere from the training domain.
This behavior almost identical for both neighborhood invariance and ATC methods (ATC-MC and ATC-NE perform the same so we report only one).
These results demonstrate that the correlation of neighborhood invariance with generalization should decrease as dataset distance increases.
To investigate the degree to which this happens in practice, we consider a set of extreme OOD experiments (Section \ref{subsec:extreme}) where we observe a surprisingly small decrease in correlation, indicating a closer dataset distance than might initially be assumed.


\begin{table}[p]
\begin{subtable}{1\textwidth}
\scriptsize
\centering
\scriptsize
\resizebox{0.65\columnwidth}{!}{%
\setlength\tabcolsep{4pt}
\begin{tabular}{lccccccc}
\toprule
& \multicolumn{3}{c}{Domain Shifts} & \multicolumn{3}{c}{ImageNet-A}\\
 \cmidrule(lr){2-4}
 \cmidrule(lr){5-7}
Measure & $R^2$ & MAE & Macro $\tau$ &  $R^2$ & MAE & Macro $\tau$ & ID $\tau$\\
\midrule
NI-RandAug & \textbf{0.709} & 11.87 & \textbf{0.724} & 0.577 & \textbf{31.17} & 0.586 & 0.845 \\
NI-Translate & 0.587 & 13.58 & 0.604 & 0.468 & 29.54 & 0.439 &  0.834 \\
NI-Erase & 0.492 & 11.86 & 0.555 & 0.446 & 33.06 & 0.517 & 0.803 \\
NI-FC & 0.603 & 12.28 & 0.691 & \textbf{0.679} & 31.84 & \textbf{0.589} & 0.867 \\
\midrule
ATC-NE & 0.607 & \textbf{11.40} & 0.703 & 0.209 & 32.00 & 0.248 & 0.931\\
ATC-MC & 0.622 & 11.97 & 0.691 & 0.159 & 32.85 & 0.190 & \textbf{0.942}\\
\bottomrule
\end{tabular}
}
\caption{Results on ImageNet scale models and datasets. On standard domain shift datasets NI-RandAug performs slightly better than ATC methods, and maintains strong performance on adversarial data where ATC methods fail completely.}
\label{tab:imagenet_corr}
\vspace{0.3cm}
\end{subtable}


\begin{subtable}{1\textwidth}
\centering
\scriptsize
\resizebox{0.9\columnwidth}{!}{%
\begin{tabular}{lccccccccc}
\toprule
& \multicolumn{5}{c}{CI10} & \multicolumn{4}{c}{Numbers}\\
 \cmidrule(lr){2-6}
 \cmidrule(lr){7-10}
Measure & $R^2$ & MAE & Macro $\tau$ & ID $\tau$ & Arch $\tau$ & $R^2$ & MAE & Macro $\tau$ & ID $\tau$ \\
\midrule
NI-RandAug & \textbf{0.899} & \textbf{3.11} & \textbf{0.768} & \textbf{0.793} & \textbf{0.837} & \textbf{0.764} & \textbf{5.33} & 0.642 & 0.733\\
NI-Translate & 0.732 & 3.56 & 0.607 & 0.661 & 0.786 & 0.685 & 6.12 & 0.667 & \textbf{0.881} \\
NI-Erase & 0.518 & 4.41 & 0.411 & 0.406 & 0.299 & 0.153 & 10.17 & -0.135 & 0.324 \\
NI-FC & 0.417 & 4.47 & 0.371 & 0.344 & 0.683 & 0.208 & 10.42 & -0.316 & -0.033\\
\midrule
ATC-NE & 0.655 & 3.61 & 0.548 & 0.693 & 0.689 & 0.616 & 6.74 & 0.637 & 0.859 \\
ATC-MC & 0.640 & 3.65 & 0.544 & 0.682 & 0.685 & 0.692 & 6.19 & \textbf{0.682} & 0.844 \\
\bottomrule
\end{tabular}
}
\caption{Results on small scale image classification, averaged across all model architectures. No Micro $\tau$ is reported since models are trained on a single domain and no Arch $\tau$ is reported for the Numbers dataset since we only consider a single architecture. NI-RandAug beats all other methods on almost all metrics.}
\label{tab:ic_corr}
\vspace{0.5cm}
\end{subtable}

\begin{subtable}{1\textwidth}
\scriptsize
\centering
\resizebox{1\columnwidth}{!}{%
\setlength\tabcolsep{4pt}
\begin{tabular}{lccccccccccc}
\toprule
& \multicolumn{5}{c}{CNN} & \multicolumn{5}{c}{RoBERTa}\\
 \cmidrule(lr){2-6}
 \cmidrule(lr){7-11}
Measure & $R^2$ & MAE & Macro $\tau$ & Micro $\tau$ & ID $\tau$ &  $R^2$ & MAE & Macro $\tau$ & Micro $\tau$ & ID $\tau$ & Arch $\tau$\\
\midrule
NI-SSMBA & 0.662 & \textbf{1.93} & \textbf{0.677} & \textbf{0.689} & \textbf{0.629} & \textbf{0.972} & \textbf{1.29} & \textbf{0.832} & \textbf{0.829} & \textbf{0.838}  & 0.588 \\
NI-EDA & 0.641 & 2.04 & 0.664  & 0.649 & 0.611 & 0.968 & 1.45 & 0.830  & 0.810 & 0.830 & 0.512\\
NI-BT & 0.550 & 2.99 & 0.592  & 0.501 & 0.538 & 0.961 & 1.47 & 0.813  & 0.801 & 0.801 & 0.523\\
NI-RandRep & 0.409 & 2.64 & 0.544 & 0.554 & 0.439 & 0.967 & 1.27 & 0.821 & 0.816 & 0.822 & 0.537 \\
\midrule
ATC-NE & 0.760 & 2.47 & 0.514 & 0.633 & 0.467 & 0.852 & 2.38 & 0.707 & 0.691 & 0.749 & 0.660 \\
ATC-MC & \textbf{0.761} & 2.46 & 0.517 & 0.634 & 0.467 & 0.869 & 2.26 & 0.722 & 0.705 & 0.749 & \textbf{0.663} \\
\bottomrule
\end{tabular}
}
\caption{Results on sentiment analysis (SA) datasets. NI-SSMBA beats all other methods on almost all metrics.}
\label{tab:sa_corr}
\vspace{0.5cm}
\end{subtable}
\begin{subtable}{1\textwidth}
\scriptsize
\centering
\resizebox{1\columnwidth}{!}{%
\setlength\tabcolsep{4pt}
\begin{tabular}{lccccccccccc}
\toprule
& \multicolumn{5}{c}{CNN} & \multicolumn{5}{c}{RoBERTa}\\
 \cmidrule(lr){2-6}
 \cmidrule(lr){7-11}
Measure & $R^2$ & MAE & Macro $\tau$ & Micro $\tau$ & ID $\tau$ &  $R^2$ & MAE & Macro $\tau$ & Micro $\tau$ & ID $\tau$ & Arch $\tau$\\
\midrule
NI-SSMBA & 0.575 & 2.09 & 0.570 & \textbf{0.534} & 0.704 & 0.933 & 1.19 & 0.750 & 0.730 & 0.771 & 0.301 \\
NI-EDA & \textbf{0.577} & \textbf{2.04} & \textbf{0.581} & 0.511 & \textbf{0.709} &  0.941 & 1.26 & \textbf{0.789} & \textbf{0.757} & \textbf{0.799} & 0.572 \\
NI-BT & 0.509 & 2.11 & 0.470 & 0.449 & 0.584 & \textbf{0.944} & \textbf{1.07} & 0.759 & 0.740 & 0.778 & 0.563 \\
NI-RandRep & 0.451 & 2.20 & 0.452 & 0.428 & 0.570 & 0.890 & 1.70 & 0.688 & 0.647 & 0.710 & 0.401 \\
\midrule
ATC-NE & 0.576 & 2.52 & 0.568 & 0.446 & 0.705 & 0.737 & 2.22 & 0.557 & 0.541 & 0.739 & 0.635\\
ATC-MC & 0.576 & 2.52 & 0.568 & 0.446 & 0.706 & 0.769 & 2.10 & 0.581 & 0.567 & 0.748 & \textbf{0.636}\\
\bottomrule
\end{tabular}
}
\caption{Results on natural language inference (NLI) datasets. NI measures beat baselines on all metrics except Arch $\tau$.}
\label{tab:nli_corr}
\vspace{0.3cm}
\end{subtable}

\caption{Evaluation metrics measuring the correlation of our neighborhood invariance measure with ID/OOD generalization.
The best performing measures for each metric are bolded.
On all tasks, neighborhood invariance achieves strong generalization and beats baseline methods on almost all metrics. Full tables for $R^2$, Macro $\tau$, Micro $\tau$, and ID $\tau$ on individual train/test domains are in Appendix \ref{app:fullres}}
\label{tab:res_full}
\end{table}


\subsection{Correlation with OOD Generalization}
We first present results in Table \ref{tab:res_full} analyzing the correlation of our proposed neighborhood invariance measure with OOD generalization.
We report $R^2$, MAE, Macro $\tau$, Micro $\tau$, ID $\tau$, and Arch $\tau$ as detailed in Section \ref{sec:metrics}.
We omit results on Spectral and Frobenius norm measures as they are close to $0$ or negative for all metrics. 
We do not report Micro $\tau$ values for image classification models since each model type is trained on only one domain. 
Additional experiments and results are presented in Appendix \ref{app:addl_experiments}.

\textbf{ImageNet-Scale Image Classification} \quad
Results on ImageNet-scale image classification datasets are presented in Table \ref{tab:imagenet_corr} and are averaged across all architectures.
On standard domain shifts, NI-RandAug performs slightly better than ATC methods on $R^2$ and Macro $\tau$ with similar MAE. 
On the adversarial ImageNet-A dataset, ATC methods fail completely whereas NI methods maintain strong performance and still correlate well with accuracy.
However, both methods exhibit large MAE and fail to accurately predict actual OOD accuracy.
On ID $\tau$ NI methods perform slightly worse than ATC methods although they still show very strong correlations.

\textbf{CIFAR10-Scale Image Classification} \quad
Results on smaller scale image classification datasets are presented in Table \ref{tab:ic_corr} and are averaged across all architectures.
On CI10 datasets, NI-RandAug significantly outperforms ATC baselines and all other measures on all metrics. 
NI-RandAug also exhibits only a small decrease in correlation when moving from in-domain (ID $\tau$) to OOD datasets (Macro $\tau$), compared to ATC methods which suffer a much larger drop.
On Numbers datasets, NI-RandAug outperforms all other methods on $R^2$ and MAE, although it performs slightly worse on Macro $\tau$ compared to NI-Translate and on ID $\tau$ compared to ATC baselines.

For both CI10 and Numbers, using patch erasing and 
flip and crop transformations cause our method to perform worse or fail entirely, in contrast to ImageNet where they perform similarly or better.
For these smaller datasets, since these transformations are more likely to 
generate images that cannot be classified (e.g. images without an object in frame) , they produce more similar invariance values between classifiers that cannot be used to rank them properly.
This effect is more pronounced for Numbers datasets because flipping the image or removing even small portions of the number to be classified can render the task impossible.
In contrast, random image translation which almost always preserves label information performs similarly well across both datasets and almost matches NI-RandAug.
We provide a larger set of ablations on the Numbers dataset exploring this phenomenon in Section \ref{sec:ablations}.
The strong performance of NI-RandAug indicates that combining multiple transformations is helpful for mitigating dataset-specific transformation sensitivities, as in the case of Numbers.


\textbf{Sentiment Analysis (SA)} \quad
Results on Sentiment Analysis datasets are presented in Table \ref{tab:sa_corr}.
In experiments on both architectures, our neighborhood invariance measures achieves strong correlation with OOD generalization and beats all baselines on almost all metrics.
Of the transformations considered, NI-SSMBA performs the best across both architectures.
NI-EDA, NI-BT, and even NI-RandRep achieve strong results as well, often beating ATC baselines.
We observe particularly strong correlation on RoBERTa models, with a nearly perfectly linear $R^2$ value of 0.972 and large Micro $\tau$ of 0.829.
We hypothesize that this is due to the pretrained initialization of RoBERTa models, which gives a strong inductive bias towards learning a space invariant to transformations that preserve meaning.
In contrast, CNN models are trained from random initializations and may not learn as closely aligned a space.
On cross architecture analysis, we observe strong Arch $\tau$ for our neighborhood measures, although they are outperformed by both ATC methods. 
Compared to image classification results, our results on sentiment analysis tasks are overall less sensitive to the data transformations selected 
because they are less likely to destroy information necessary for classification.

\textbf{Natural Language Inference (NLI)} \quad
Results on Natural Language Inference tasks are presented in Table \ref{tab:nli_corr}.
Similar to our sentiment analysis results, our neighborhood invariance measures achieve strong correlation with OOD generalization on both architectures and beat all baselines.
Correlations in general on NLI are lower than those of sentiment analysis because it is more difficult to maintain the complex relationship between the two sentences during a transformation.
For example, changing a single word can easily change a sentence pair from entailment to contradiction, whereas many words must be changed to modify a 5 star review to a 1 star review.
We observe exceptionally high correlation on RoBERTa models, for which we offer a similar hypothesis as in our sentiment analysis experiments.
On cross architecture analysis, we observe strong correlation for our NI-EDA and NI-BT although they are outperformed by both ATC methods.

\subsection{Extreme OOD Generalization}
\label{subsec:extreme}

\begin{table}
\begin{subtable}{0.465\textwidth}
\centering
\small
\resizebox{1\columnwidth}{!}{%

\begin{tabular}{lcccc}
\toprule
& \multicolumn{2}{c}{CNN} & \multicolumn{2}{c}{RoBERTa}\\
 \cmidrule(lr){2-3}
 \cmidrule(lr){4-5}
Measure & $R^2$ & Micro $\tau$ & $R^2$ & Micro $\tau$\\
\midrule
NI-SSMBA & \textbf{0.584} & \textbf{0.566} & \textbf{0.941} & \textbf{0.816} \\
NI-EDA & 0.575 & \textbf{0.567} & 0.884 & 0.715 \\
NI-BT & 0.538 & 0.470 & 0.906 & 0.766 \\
NI-RandRep & 0.277 & 0.373 & 0.918 & 0.776 \\
\midrule
ATC-NE & 0.271 & 0.495 & 0.329 & 0.356  \\
ATC-MC & 0.295 & 0.506 & 0.437 & 0.436 \\
\bottomrule
\end{tabular}
}
\caption{Results on the \texttt{Drugs.com} dataset.}
\label{tab:sa_extreme}
\end{subtable}
\hfill
\begin{subtable}{0.465\textwidth}
\centering
\small
\resizebox{1\columnwidth}{!}{%
\begin{tabular}{lcccc}
\toprule
& \multicolumn{2}{c}{CNN} & \multicolumn{2}{c}{RoBERTa}\\
 \cmidrule(lr){2-3}
 \cmidrule(lr){4-5}
Measure & $R^2$ & Micro $\tau$ & $R^2$ & Micro $\tau$\\
\midrule
NI-SSMBA & 0.083 & 0.080 & 0.691 & 0.463 \\
NI-EDA & 0.202 & 0.110 & \textbf{0.739} & \textbf{0.540} \\
NI-BT & \textbf{0.213} & \textbf{0.247} & 0.730 & 0.527 \\
NI-RandRep & 0.096 & 0.102 & 0.030 & 0.012 \\
\midrule
ATC-NE & 0.077 & -0.107 & 0.719 & 0.345 \\
ATC-MC & 0.076 & -0.106 & 0.734 & 0.354 \\
\bottomrule
\end{tabular}
}
\caption{Results on the MedNLI dataset.}
\label{tab:nli_extreme}
\end{subtable}
\caption{Evaluation metrics measuring the correlation of 
our neighborhood invariance measure with generalization on extreme OOD datasets. 
NI-* methods beat baselines on both tasks, with
RoBERTa models exhibiting only a slight degradation in correlation compared to more typical OOD settings.}
\label{tab:extreme}
\vspace{0.3cm}
\end{table}

We now consider more extreme generalization to data domains with specialized and knowledge intensive data.
We consider only natural language tasks as it is difficult to find a sufficiently specialized image classification dataset that maintains the same output classes.
For sentiment analysis we use the \texttt{Drugs.com} review dataset \citep{grasser2018aspect}, and for natural language inference we use MedNLI \citep{romanov2018lessons}, an NLI dataset generated from clinical notes and patient history.
Both datasets contain highly specific medical language not seen in any of our training domains.
All models from all original training domains are evaluated on each of these extreme OOD domains, and we report $R^2$ and Micro $\tau$. 
Results are shown in Table \ref{tab:extreme}

On the \texttt{Drugs.com} dataset, we observe a small decrease in correlation for neighborhood invariance methods compared to results on AWS datasets. 
However, ATC methods begin to fail, with Micro $\tau$ on RoBERTa models dropping significantly from 0.706 to 0.356.
This suggests that models become poorly calibrated in extreme OOD settings, making ATC methods fragile.
On MedNLI we observe a much larger disparity in performance. 
For CNN models, most of our measures fail to correlate at all, and ATC methods degrade so much they became anti correlated with generalization.
For RoBERTa models we observe only minor drops in correlation for all measures.
For both tasks NI-RandRep exhibits almost no correlation with OOD generalization. 
This suggests that the choice of transformation becomes much more important as we move farther from our training domain.

\subsection{Ablations}
\label{sec:ablations}
In this section we examine factors that may affect the quality of neighborhood invariance estimation and its correlation with actual generalization.
Since rerunning all of our experiments is too costly, we evaluate on \texttt{toys} Amazon reviews using a pool of CNN models trained on all other domains for the first three ablations, and on the Numbers datasets using NiN models trained on SVHN for the final two. 

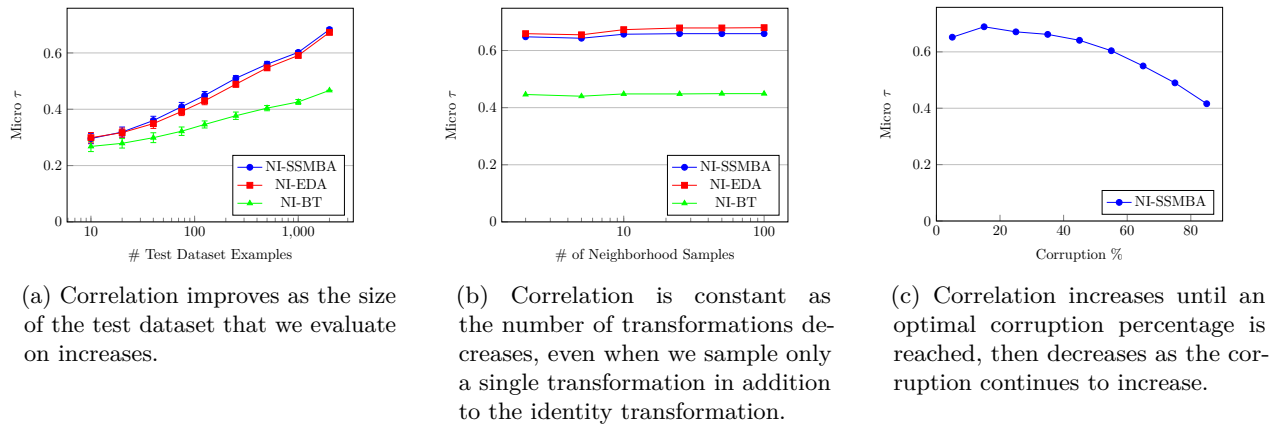
\begin{figure}
     \centering
     \begin{subfigure}[t]{0.3\textwidth}
\centering
\begin{tikzpicture}[scale=0.55, trim axis left]
\begin{axis}[
    xmode=log,
    log ticks with fixed point,
    xlabel=\# Test Dataset Examples,
    ylabel=Micro $\tau$,
    legend pos=south east,
    xtick pos=left,
    ytick pos=left,
    ymin=0,
    ymajorgrids=true,
    height=6.75cm, width=8.5cm,
]

\addplot+[blue, mark options={blue},error bars/.cd, y dir=both, y explicit,] table [x=x, y=y,y error=error, col sep=comma] {
    x,  y, error
10, 0.296,  0.018324702393432653
20, 0.319, 0.01791421085657947
40, 0.360, 0.015050447656614967
75, 0.409, 0.015255267279411852
125, 0.449, 0.014111996466274602
250, 0.510, 0.009548011949722242
500, 0.560, 0.009973199354404164
1000, 0.602, 0.006746716538449564
2000, 0.683
};
\addlegendentry{NI-SSMBA}

\addplot+[red, mark options={red},error bars/.cd, y dir=both, y explicit,] table [x=x, y=y,y error=error, col sep=comma] {
    x,  y, error
10, 0.299, 0.0184
20, 0.317, 0.016
40, 0.349, 0.017
75, 0.392, 0.0145
125, 0.430, 0.0139
250, 0.489, 0.0106
500, 0.547, 0.0092
1000, 0.591, 0.0064
2000, 0.673
};
\addlegendentry{NI-EDA}

\addplot+[green, mark options={green},error bars/.cd, y dir=both, y explicit,] table [x=x, y=y,y error=error, col sep=comma] {
    x,  y, error
10, 0.268, 0.0183
20, 0.279, 0.0168
40, 0.299, 0.0175
75, 0.322, 0.015
125, 0.346, 0.0125
250, 0.377, 0.0131
500, 0.404, 0.0093
1000, 0.426, 0.0087
2000, 0.467, 0
};
\addlegendentry{NI-BT}


\end{axis}
\end{tikzpicture}
\caption{Correlation improves as the size of the test dataset that we evaluate on increases.}
\label{fig:dset_exp}
     \end{subfigure}
     \hfill
     \begin{subfigure}[t]{0.3\textwidth}
\centering
\begin{tikzpicture}[scale=0.55, trim axis left]
\begin{axis}[
    xmode=log,
    log ticks with fixed point,
    xlabel=\# of Neighborhood Samples,
    ylabel=Micro $\tau$,
    legend pos=south east,
    xtick pos=left,
    ytick pos=left,
    ymajorgrids=true,
    height=6.75cm, width=8.5cm,
    ymin=0,
]

\addplot+[blue, mark options={blue}] table [x=x, y=y,y error=error, col sep=comma] {
    x,  y, error
2, 0.648, 0.002245408079183461
5, 0.643, 0.0013726153990928336
10, 0.657, 0.0009442264918753964
25, 0.659, 0.0006928922945864645
50, 0.659, 0.0003973057444118493
100, 0.659, 1.1102230246251565e-16
};
\addlegendentry{NI-SSMBA}

\addplot+[red, mark options={red}] table [x=x, y=y,y error=error, col sep=comma] {
    x,  y
2, 0.659
5, 0.655
10, 0.673
25, 0.679
50, 0.679
100, 0.680
};
\addlegendentry{NI-EDA}

\addplot+[green, mark options={green}] table [x=x, y=y,y error=error, col sep=comma] {
    x,  y
2, 0.446
5, 0.440
10, 0.448
25, 0.448
50, 0.449
100, 0.449
};
\addlegendentry{NI-BT}

\end{axis}
\end{tikzpicture}
\caption{Correlation is constant as the number of transformations decreases, even when we sample only a single transformation in addition to the identity transformation.}
\label{fig:num_generated}
     \end{subfigure}
     \hfill
     \begin{subfigure}[t]{0.3\textwidth}
\centering
\begin{tikzpicture}[scale=0.55, trim axis left]
\begin{axis}[
    xlabel=Corruption \%,
    ylabel=Micro $\tau$,
    xtick pos=left,
    ytick pos=left,
    ymajorgrids=true,
    xmin=0,xmax=90,
    ymin=0,
    legend pos=south east,
    height=6.75cm, width=8.5cm,
]


\addplot+[blue, mark options={blue}] table [x=x, y=y, col sep=comma] {
x,  y
5, 0.652
15, 0.689
25, 0.671
35, 0.662
45, 0.641
55, 0.604
65, 0.550
75, 0.490
85, 0.416
};
\addlegendentry{NI-SSMBA}

\end{axis}
\end{tikzpicture}
\caption{Correlation increases until an optimal corruption percentage is reached, then decreases as the corruption continues to increase.}
\label{fig:neighborhood_size}
\end{subfigure}
\caption{Micro $\tau$ for our neighborhood invariance measure calculated with varying ablations on CNN models evaluated on Amazon \texttt{toy} reviews.}
\label{fig:ablations}
\end{figure}

\textbf{Test Dataset Size:} 
Does our neighborhood invariance measure still correlate well when the test dataset is small? 
We randomly and iteratively subsample our test dataset of 2000 examples to reduce our dataset size down to 10 examples.
We then measure our models' neighborhood invariance on each subsampled dataset and calculate the Micro $\tau$ on all models.
Results are shown in Figure \ref{fig:dset_exp}.
We find that for all neighborhoods, smaller datasets lead to noisier invariance estimates and lower correlation.
As dataset size increases, correlation increases as well.

\textbf{Number of Transformations:}
How many transformations do we need to sample in order to generate a reliable neighborhood invariance estimate?
We sample a varying number of transformations for each test example, from a minimum of two transformations to a maximum of 100 transformations, then estimate our neighborhood invariance measure with each set of transformations on the entire test dataset and calculate the micro $\tau$.
By default we always include the identity transformation.
Results are shown in Figure \ref{fig:num_generated}.
We find that our measure is surprisingly robust to the number of samples, with only a small difference between 100 and 2 transformations sampled.
For all measures, correlation slightly increases as the number of samples increases and we achieve a better estimation of the true invariance value.

\textbf{Transformation Magnitude:} 
How sensitive is our measure to the maximum magnitude of transformation considered?
We use the SSMBA transformation for which the magnitude of a transformation is defined by the percentage of tokens corrupted, which we vary from a minimum of 5\% to a maximum of 85\%.
After sampling a set of transformations from each corruption level, we estimate neighborhood invariance on the test dataset using each set 
and calculate the micro $\tau$ on all models.
Results are shown in Figure \ref{fig:neighborhood_size}.
We find that as we begin to increase the corruption percentage, correlation begins to increase as well.
Correlation reaches a maximum, then decreases as we continue to increase our corruption percentage.
However, even at 85\% corruption, 
our method is quite robust and achieves a 
micro $\tau$ of 0.416.

\textbf{Selecting Transformations:}
How do we ensure that transformations are suitable for a given dataset and do not destroy label information?
We consider the set of NiN models trained on SVHN and a set of image transformations including RandAugment \citep{cubuk2020randaugment}, rotations, translations, shears, brightness jittering, contrast jittering, color jittering, patch erasing, and flips and crops.
For each transformation and both OOD datasets ColoredMNIST and MNIST we calculate the Macro $\tau$ correlation between accuracy and neighborhood invariance, as well as the average entropy difference between a model's output on a transformed image and on the original image.
Results are shown in Figure \ref{fig:tablation}.

Neighborhood invariance is relatively insensitive to the transformation selected, with most transformations performing similarly up to a certain entropy difference threshold around 0.1, after which it fails. 
The transformations that fail, erase and flip crop, both tend to destroy label information and lead to much higher entropy outputs.
We propose this method of examining entropy differences as a simple way to diagnose whether a given transform is appropriate for a specific dataset.

\textbf{Pretraining and Training with Augmentation:}
Does self-supervised pretraining or training with data augmentations, 
which should make models more invariant to certain transformations, make our neighborhood invariance measure ineffective?
We begin by examining our metrics on subsets of models from the ImageNet Testbed \citep{taori2020measuring} split by models trained on standard ImageNet (81 models), models trained on ImageNet with augmentations and robustness interventions (74 models), and finally models trained with extra data or pretrained with self-supervised objectives (41 models).
Results are shown in Table \ref{tab:imagenet_ablations}.
We find that compared to evaluating only on standard training models, the addition of augmentations or pretraining slightly degrades Macro $\tau$, but improves $R^2$ and MAE. Compared to the overall results on all models, we do not observe any large decreases in performance.

Since the augmentations considered in the set of ImageNet Testbed models are not the same across models, we also 
consider models from PGDL \citep{jiang2020predicting} trained with and without a single type of augmentation: flip crop.
Calculating our evaluation metrics on each set of models allows us to isolate the effect of data augmentation.
Results are shown in Table \ref{app:tab_aug}.
We find that measuring neighborhood invariance using the same transformation (NI-FC) that models are trained with causes only a slight degradation compared to models trained without.
When measuring invariance using other transformations (NI-RandAug, NI-Erase), evaluation metrics actually improve slightly.

\begin{table}
    \centering
    \small
    \begin{tabular}{lcccc}
\toprule
Training Regime & $R^2$ & MAE & Macro $\tau$ & ID $\tau$ \\
\midrule
Standard Training & 0.684 & 12.37 & \textbf{0.763} & 0.852 \\
+ Augmentation/Robustness & 0.707 & 12.23 & 0.650 & \textbf{0.855} \\
+ Pretraining/Extra Data & \textbf{0.799} & \textbf{11.73} & 0.707 & 0.836 \\
\midrule
All Models & 0.709 & 11.87 & 0.724 & 0.845 \\
\bottomrule
    \end{tabular}
    \caption{NI-RandAug results on different subsets of the ImageNet Testbed models trained with different training regimes. Training with augmentation and robustness interventions decreases macro $\tau$ compared to standard training but increases all other metrics. $R^2$ and MAE are higher for models trained/pretrained with large amounts of data.}
    \label{tab:imagenet_ablations}
\end{table}

\begin{figure}
    \begin{minipage}[h]{0.35\textwidth}
    \centering
    \begin{tikzpicture}[scale=0.55]
    \begin{axis}[
        xlabel=Average Entropy $\Delta$,
        ylabel=Macro $\tau$,
        xtick pos=left,
        ytick pos=left,
        ymajorgrids=true,
        xmin=-0.1,xmax=0.3,
        ymin=-0.5,ymax=1,
        legend pos=outer north east,
        height=6.75cm, width=8.5cm,
    ]
    
    
    \addplot+[only marks, mark=*] table [x=x, y=y, col sep=comma] {
    y, x
    0.6682978387956844, 0.0432478
    0.6142557651991614, 0.056486376
    };
    \addlegendentry{RandAugment}

    \addplot+[only marks, mark=*] table [x=x, y=y, col sep=comma] {
    y, x
    0.7591751599708296, 0.06016565
    0.6925227113906358, 0.06275674
    };
    \addlegendentry{Affine 10}

    \addplot+[only marks, mark=*] table [x=x, y=y, col sep=comma] {
    y, x
    -0.09646977170900045, 0.14383769
    -0.16561844863731653, 0.15311976
    };
    \addlegendentry{Erase}

    \addplot+[only marks, mark=*] table [x=x, y=y, col sep=comma] {
    y, x
    -0.20831878238610244, 0.12754895
    -0.3878406708595387, 0.13490444
    };
    \addlegendentry{FlipCrop}

    \addplot+[only marks, mark=*] table [x=x, y=y, col sep=comma] {
    y, x
    0.7284166820346266, 0.057083193
    0.758211041229909, 0.053090964
    };
    \addlegendentry{Rotate 20}

    \addplot+[only marks, mark=*] table [x=x, y=y, col sep=comma] {
    y, x
    0.7228242315007715, 0.08633844
    0.7624039133473095, 0.08321828
    };
    \addlegendentry{Rotate 30}

    \addplot+[only marks, mark=*] table [x=x, y=y, col sep=comma] {
    y, x
    0.6864733030307133, 0.031048667
    0.6184486373165619, 0.03517614
    };
    \addlegendentry{Translate 10}

    \addplot+[only marks, mark=*] table [x=x, y=y, col sep=comma] {
    y, x
    -0.099265996975928, 0.14734809
-0.20335429769392033, 0.15018375
    };
    \addlegendentry{Translate 30}
    
    \addplot+[only marks, mark=*] table [x=x, y=y, col sep=comma] {
    y, x
    0.7242223441342353, 0.044073734
    0.7721872816212438, 0.04087309
    };
    \addlegendentry{Shear 15}
    
    \addplot+[only marks, mark=*] table [x=x, y=y, col sep=comma] {
    y, x
    0.7442348008385743, -0.009599328
    };
    \addlegendentry{Color}
    
    \addplot+[only marks, mark=*] table [x=x, y=y, col sep=comma] {
    y, x
    0.5830129681543941, -0.0011068167
    0.6981132075471698, 0.004066915
    };
    \addlegendentry{Contrast}

    \addplot+[only marks, mark=*] table [x=x, y=y, col sep=comma] {
    y, x
    0.5816148555209303, 0.0013248248
0.710691823899371, 0.0067206305
    };
    \addlegendentry{Brightness}

    \end{axis}
    \end{tikzpicture}
    \caption{Transformations whose neighborhood invariance measures correlate well with OOD generalization exhibit smaller differences in output entropy.}
    \label{fig:tablation}
    \end{minipage}
    \hfill
    \begin{minipage}[h]{0.6\textwidth}
        \centering
        \footnotesize
        \begin{tabular}{lcccccc}
        \toprule
        Measure & Flip Crop & $R^2$ & MAE & Macro $\tau$ & ID $\tau$\\
        \midrule
        \multirow{2}{*}{NI-RandAug} & \xmark & 0.833 & 3.32 & 0.719 & \textbf{0.753} \\
        & \checkmark & \textbf{0.844} & \textbf{3.16} & \textbf{0.732} & 0.744 \\
        \midrule
        \multirow{2}{*}{NI-Translate} & \xmark & 0.874 & 2.61 & 0.768 & 0.831 \\
        & \checkmark & \textbf{0.877} & \textbf{2.52} & \textbf{0.794} & \textbf{0.845} \\
        \midrule
        \multirow{2}{*}{NI-Erase} &\xmark  & 0.790 & 3.28 & 0.716 & 0.702\\
        & \checkmark & \textbf{0.812} & \textbf{3.17} & \textbf{0.722} & \textbf{0.719}\\
        \midrule
        \multirow{2}{*}{NI-FC} & \xmark & \textbf{0.681} & 4.34 & \textbf{0.620} & \textbf{0.587} \\
        & \checkmark & 0.663 & \textbf{4.23} & 0.608 & 0.554\\
        \bottomrule
        \end{tabular}
        \captionof{table}{Training with and without flip crop augmentation has a minimal effect on the effectiveness of our method, even when the transformation neighborhood aligns with those used to train the model (NI-FC).}
        \label{app:tab_aug}
    \end{minipage}
\end{figure}



\section{Discussion}
\label{sec:discussion}
In this paper, motivated by the limited settings in which existing complexity measures can be applied, we propose a simple to calculate neighborhood invariance measure that can be applied even when test distributions are unknown and model training data, weights, and gradients are unavailable.
We evaluate our method on image classification, sentiment analysis, and natural language inference datasets, calculating a variety of correlation metrics with both in-domain and out-of-domain (OOD) generalization.
Across almost all tasks and experimental settings, we 
find that our neighborhood invariance measure consistently outperforms baseline methods and correlates strongly with actual generalization.
However, our method has several limitations. 
Data transformations must be selected such that labels for  transformed points are still well defined, although examining entropy differences can diagnose poor transformation choices
In settings where such transformations are difficult to define, our method may not be applicable or provide inappropriately high estimates, so practitioners must be careful to verify their estimates with a labelled test set.
In addition, our neighborhood invariance measure may fail in sufficiently OOD settings where a model may become poorly calibrated or degenerate, although we find in practice on our tasks that even extreme OOD settings are similar enough for our measure to perform well.
In future work we plan to explore using similar measures calculated over transformation neighborhoods as a method for OOD detection.

\section{Acknowledgments}
We would like to thank Taylor Killian, Tom Hartvigsen, and Swami Sankaranarayanan for their helpful discussion and comments.
Resources used in preparing this research were provided, in part, by the Province of Ontario, the Government of Canada through CIFAR, and companies sponsoring the Vector Institute (\url{www.vectorinstitute.ai/partners}).

\bibliography{main}

\begin{thebibliography}{106}
\providecommand{\natexlab}[1]{#1}
\providecommand{\url}[1]{\texttt{#1}}
\expandafter\ifx\csname urlstyle\endcsname\relax
  \providecommand{\doi}[1]{doi: #1}\else
  \providecommand{\doi}{doi: \begingroup \urlstyle{rm}\Url}\fi

\bibitem[{Aithal K} et~al.(2021){Aithal K}, {Kashyap}, and
  {Subramanyam}]{aithal2021robustness}
Sumukh {Aithal K}, Dhruva {Kashyap}, and Natarajan {Subramanyam}.
\newblock {Robustness to Augmentations as a Generalization metric}.
\newblock \emph{arXiv e-prints}, art. arXiv:2101.06459, January 2021.

\bibitem[{Allen-Zhu} et~al.(2018){Allen-Zhu}, {Li}, and
  {Liang}]{zhu2018learning}
Zeyuan {Allen-Zhu}, Yuanzhi {Li}, and Yingyu {Liang}.
\newblock {Learning and Generalization in Overparameterized Neural Networks,
  Going Beyond Two Layers}.
\newblock \emph{arXiv e-prints}, art. arXiv:1811.04918, November 2018.

\bibitem[{Anselmi} et~al.(2015){Anselmi}, {Rosasco}, and
  {Poggio}]{anselmi2015invariance}
Fabio {Anselmi}, Lorenzo {Rosasco}, and Tomaso {Poggio}.
\newblock {On Invariance and Selectivity in Representation Learning}.
\newblock \emph{arXiv e-prints}, art. arXiv:1503.05938, March 2015.

\bibitem[Anselmi et~al.(2016)Anselmi, Leibo, Rosasco, Mutch, Tacchetti, and
  Poggio]{ANSELMI2016112}
Fabio Anselmi, Joel~Z. Leibo, Lorenzo Rosasco, Jim Mutch, Andrea Tacchetti, and
  Tomaso Poggio.
\newblock Unsupervised learning of invariant representations.
\newblock \emph{Theoretical Computer Science}, 633:\penalty0 112--121, 2016.
\newblock ISSN 0304-3975.
\newblock \doi{https://doi.org/10.1016/j.tcs.2015.06.048}.
\newblock URL
  \url{https://www.sciencedirect.com/science/article/pii/S0304397515005587}.
\newblock Biologically Inspired Processes in Neural Computation.

\bibitem[Arjovsky et~al.(2019)Arjovsky, Bottou, Gulrajani, and
  Lopez-Paz]{InvariantRiskMinimization}
Martin Arjovsky, L{\'e}on Bottou, Ishaan Gulrajani, and David Lopez-Paz.
\newblock Invariant risk minimization.
\newblock \emph{arXiv}, 2019.

\bibitem[Baek et~al.(2022)Baek, Jiang, Raghunathan, and
  Kolter]{baek2022agreementontheline}
Christina Baek, Yiding Jiang, Aditi Raghunathan, and Zico Kolter.
\newblock Agreement-on-the-line: Predicting the performance of neural networks
  under distribution shift, 2022.

\bibitem[Barbu et~al.(2019)Barbu, Mayo, Alverio, Luo, Wang, Gutfreund,
  Tenenbaum, and Katz]{barbu2019objectnet}
Andrei Barbu, David Mayo, Julian Alverio, William Luo, Christopher Wang, Dan
  Gutfreund, Josh Tenenbaum, and Boris Katz.
\newblock Objectnet: A large-scale bias-controlled dataset for pushing the
  limits of object recognition models.
\newblock In H.~Wallach, H.~Larochelle, A.~Beygelzimer, F.~d\textquotesingle
  Alch\'{e}-Buc, E.~Fox, and R.~Garnett (eds.), \emph{Advances in Neural
  Information Processing Systems}, volume~32. Curran Associates, Inc., 2019.
\newblock URL
  \url{https://proceedings.neurips.cc/paper_files/paper/2019/file/97af07a14cacba681feacf3012730892-Paper.pdf}.

\bibitem[Bartlett \& Mendelson(2003)Bartlett and
  Mendelson]{bartlett2003rademacher}
Peter~L. Bartlett and Shahar Mendelson.
\newblock Rademacher and gaussian complexities: Risk bounds and structural
  results.
\newblock \emph{J. Mach. Learn. Res.}, 3\penalty0 (null):\penalty0 463–482,
  mar 2003.
\newblock ISSN 1532-4435.

\bibitem[Bartlett et~al.(2017)Bartlett, Foster, and
  Telgarsky]{bartlett2017spectrally}
Peter~L Bartlett, Dylan~J Foster, and Matus~J Telgarsky.
\newblock Spectrally-normalized margin bounds for neural networks.
\newblock In I.~Guyon, U.~V. Luxburg, S.~Bengio, H.~Wallach, R.~Fergus,
  S.~Vishwanathan, and R.~Garnett (eds.), \emph{Advances in Neural Information
  Processing Systems}, volume~30. Curran Associates, Inc., 2017.
\newblock URL
  \url{https://proceedings.neurips.cc/paper/2017/file/b22b257ad0519d4500539da3c8bcf4dd-Paper.pdf}.

\bibitem[Ben-David et~al.(2007)Ben-David, Blitzer, Crammer, and
  Pereira]{bendavid2007analysis}
Shai Ben-David, John Blitzer, Koby Crammer, and Fernando Pereira.
\newblock Analysis of representations for domain adaptation.
\newblock In B.~Sch\"{o}lkopf, J.~Platt, and T.~Hoffman (eds.), \emph{Advances
  in Neural Information Processing Systems}, volume~19. MIT Press, 2007.
\newblock URL
  \url{https://proceedings.neurips.cc/paper/2006/file/b1b0432ceafb0ce714426e9114852ac7-Paper.pdf}.

\bibitem[Beyer et~al.(2021)Beyer, Zhai, Royer, Markeeva, Anil, and
  Kolesnikov]{beyer2021bit}
Lucas Beyer, Xiaohua Zhai, Am{\'{e}}lie Royer, Larisa Markeeva, Rohan Anil, and
  Alexander Kolesnikov.
\newblock Knowledge distillation: {A} good teacher is patient and consistent.
\newblock \emph{CoRR}, abs/2106.05237, 2021.
\newblock URL \url{https://arxiv.org/abs/2106.05237}.

\bibitem[Brown et~al.(2020)Brown, Mann, Ryder, Subbiah, Kaplan, Dhariwal,
  Neelakantan, Shyam, Sastry, Askell, Agarwal, Herbert-Voss, Krueger, Henighan,
  Child, Ramesh, Ziegler, Wu, Winter, Hesse, Chen, Sigler, Litwin, Gray, Chess,
  Clark, Berner, McCandlish, Radford, Sutskever, and Amodei]{brown2020gpt3}
Tom Brown, Benjamin Mann, Nick Ryder, Melanie Subbiah, Jared~D Kaplan, Prafulla
  Dhariwal, Arvind Neelakantan, Pranav Shyam, Girish Sastry, Amanda Askell,
  Sandhini Agarwal, Ariel Herbert-Voss, Gretchen Krueger, Tom Henighan, Rewon
  Child, Aditya Ramesh, Daniel Ziegler, Jeffrey Wu, Clemens Winter, Chris
  Hesse, Mark Chen, Eric Sigler, Mateusz Litwin, Scott Gray, Benjamin Chess,
  Jack Clark, Christopher Berner, Sam McCandlish, Alec Radford, Ilya Sutskever,
  and Dario Amodei.
\newblock Language models are few-shot learners.
\newblock In H.~Larochelle, M.~Ranzato, R.~Hadsell, M.F. Balcan, and H.~Lin
  (eds.), \emph{Advances in Neural Information Processing Systems}, volume~33,
  pp.\  1877--1901. Curran Associates, Inc., 2020.
\newblock URL
  \url{https://proceedings.neurips.cc/paper_files/paper/2020/file/1457c0d6bfcb4967418bfb8ac142f64a-Paper.pdf}.

\bibitem[Bühlmann(2018)]{bühlmann2018invariance}
Peter Bühlmann.
\newblock Invariance, causality and robustness, 2018.

\bibitem[Chen* et~al.(2021)Chen*, Goel*, Sohoni*, Poms, Fatahalian, and
  Re]{chenmandoline2021}
Mayee Chen*, Karan Goel*, Nimit Sohoni*, Fait Poms, Kayvon Fatahalian, and
  Christopher Re.
\newblock Mandoline: Model evaluation under distribution shift.
\newblock \emph{International Conference of Machine Learning (ICML)}, 2021.

\bibitem[{Chen} et~al.(2019){Chen}, {Jiang}, {Liao}, and
  {Zhao}]{chen2019nonparametric}
Minshuo {Chen}, Haoming {Jiang}, Wenjing {Liao}, and Tuo {Zhao}.
\newblock {Nonparametric Regression on Low-Dimensional Manifolds using Deep
  ReLU Networks : Function Approximation and Statistical Recovery}.
\newblock \emph{arXiv e-prints}, art. arXiv:1908.01842, August 2019.

\bibitem[Chuang et~al.(2020)Chuang, Torralba, and
  Jegelka]{chuang2020estimating}
Ching-Yao Chuang, Antonio Torralba, and Stefanie Jegelka.
\newblock Estimating generalization under distribution shifts via
  domain-invariant representations.
\newblock \emph{International conference on machine learning}, 2020.

\bibitem[Cubuk et~al.(2020)Cubuk, Zoph, Shlens, and Le]{cubuk2020randaugment}
Ekin~Dogus Cubuk, Barret Zoph, Jon Shlens, and Quoc Le.
\newblock Randaugment: Practical automated data augmentation with a reduced
  search space.
\newblock In H.~Larochelle, M.~Ranzato, R.~Hadsell, M.F. Balcan, and H.~Lin
  (eds.), \emph{Advances in Neural Information Processing Systems}, volume~33,
  pp.\  18613--18624. Curran Associates, Inc., 2020.
\newblock URL
  \url{https://proceedings.neurips.cc/paper/2020/file/d85b63ef0ccb114d0a3bb7b7d808028f-Paper.pdf}.

\bibitem[{Darlow} et~al.(2018){Darlow}, {Crowley}, {Antoniou}, and
  {Storkey}]{darlow2018cinic10}
Luke~N. {Darlow}, Elliot~J. {Crowley}, Antreas {Antoniou}, and Amos~J.
  {Storkey}.
\newblock {CINIC-10 is not ImageNet or CIFAR-10}.
\newblock \emph{arXiv e-prints}, art. arXiv:1810.03505, October 2018.

\bibitem[Deng et~al.(2009)Deng, Dong, Socher, Li, Li, and
  Fei-Fei]{deng2009imagenet}
Jia Deng, Wei Dong, Richard Socher, Li-Jia Li, Kai Li, and Li~Fei-Fei.
\newblock Imagenet: A large-scale hierarchical image database.
\newblock In \emph{2009 IEEE Conference on Computer Vision and Pattern
  Recognition}, pp.\  248--255, 2009.
\newblock \doi{10.1109/CVPR.2009.5206848}.

\bibitem[Deng(2012)]{deng2012mnist}
Li~Deng.
\newblock The mnist database of handwritten digit images for machine learning
  research.
\newblock \emph{IEEE Signal Processing Magazine}, 29\penalty0 (6):\penalty0
  141--142, 2012.

\bibitem[Deng \& Zheng(2021)Deng and Zheng]{deng2020labels}
Weijian Deng and Liang Zheng.
\newblock Are labels always necessary for classifier accuracy evaluation?
\newblock In \emph{Proc. CVPR}, 2021.

\bibitem[Deng et~al.(2021)Deng, Gould, and Zheng]{Deng:ICML2021}
Weijian Deng, Stephen Gould, and Liang Zheng.
\newblock What does rotation prediction tell us about classifier accuracy under
  varying testing environments?
\newblock In \emph{ICML}, 2021.

\bibitem[Dosovitskiy et~al.(2020)Dosovitskiy, Beyer, Kolesnikov, Weissenborn,
  Zhai, Unterthiner, Dehghani, Minderer, Heigold, Gelly, Uszkoreit, and
  Houlsby]{dosovitskiy2020vit}
Alexey Dosovitskiy, Lucas Beyer, Alexander Kolesnikov, Dirk Weissenborn,
  Xiaohua Zhai, Thomas Unterthiner, Mostafa Dehghani, Matthias Minderer, Georg
  Heigold, Sylvain Gelly, Jakob Uszkoreit, and Neil Houlsby.
\newblock An image is worth 16x16 words: Transformers for image recognition at
  scale.
\newblock In \emph{International Conference on Learning Representations}, 2020.

\bibitem[Dziugaite \& Roy(2017)Dziugaite and Roy]{dziugaite2017computing}
Gintare~Karolina Dziugaite and Daniel~M. Roy.
\newblock Computing nonvacuous generalization bounds for deep (stochastic)
  neural networks with many more parameters than training data.
\newblock In \emph{Proceedings of the 33rd Annual Conference on Uncertainty in
  Artificial Intelligence (UAI)}, 2017.

\bibitem[{Garg} et~al.(2022){Garg}, {Balakrishnan}, {Lipton}, {Neyshabur}, and
  {Sedghi}]{2022arXiv220104234G}
Saurabh {Garg}, Sivaraman {Balakrishnan}, Zachary~C. {Lipton}, Behnam
  {Neyshabur}, and Hanie {Sedghi}.
\newblock {Leveraging Unlabeled Data to Predict Out-of-Distribution
  Performance}.
\newblock \emph{arXiv e-prints}, art. arXiv:2201.04234, January 2022.

\bibitem[Garg et~al.(2021)Garg, Tauman~Kalai, Ligett, and
  Wu]{pmlr-v130-garg21a}
Vikas Garg, Adam Tauman~Kalai, Katrina Ligett, and Steven Wu.
\newblock Learn to expect the unexpected: Probably approximately correct domain
  generalization.
\newblock In Arindam Banerjee and Kenji Fukumizu (eds.), \emph{Proceedings of
  The 24th International Conference on Artificial Intelligence and Statistics},
  volume 130 of \emph{Proceedings of Machine Learning Research}, pp.\
  3574--3582. PMLR, 13--15 Apr 2021.
\newblock URL \url{https://proceedings.mlr.press/v130/garg21a.html}.

\bibitem[Glantz et~al.(1990)Glantz, Slinker, and Neilands]{glantz1990primer}
Stanton~A Glantz, Bryan~K Slinker, and Torsten~B Neilands.
\newblock \emph{Primer of Applied Regression and Analysis of Variance}.
\newblock Health Professions Division, McGraw-Hill, New York, 1990.

\bibitem[Goodfellow et~al.(2014)Goodfellow, Shlens, and
  Szegedy]{goodfellow2014explaining}
Ian Goodfellow, Jonathon Shlens, and Christian Szegedy.
\newblock Explaining and harnessing adversarial examples.
\newblock \emph{arXiv 1412.6572}, 12 2014.

\bibitem[Gr\"{a}\ss{}er et~al.(2018)Gr\"{a}\ss{}er, Kallumadi, Malberg, and
  Zaunseder]{grasser2018aspect}
Felix Gr\"{a}\ss{}er, Surya Kallumadi, Hagen Malberg, and Sebastian Zaunseder.
\newblock Aspect-based sentiment analysis of drug reviews applying cross-domain
  and cross-data learning.
\newblock In \emph{Proceedings of the 2018 International Conference on Digital
  Health}, DH '18, pp.\  121–125, New York, NY, USA, 2018. Association for
  Computing Machinery.
\newblock ISBN 9781450364935.
\newblock \doi{10.1145/3194658.3194677}.
\newblock URL \url{https://doi.org/10.1145/3194658.3194677}.

\bibitem[Gu et~al.(2019)Gu, Yang, Ngiam, Le, and Shlens]{gu2019using}
Keren Gu, Brandon Yang, Jiquan Ngiam, Quoc Le, and Jonathon Shlens.
\newblock Using videos to evaluate image model robustness.
\newblock In \emph{SafeML Workshop at ICLR}, 2019.

\bibitem[{Guillory} et~al.(2021){Guillory}, {Shankar}, {Ebrahimi}, {Darrell},
  and {Schmidt}]{guillory2021predicting}
Devin {Guillory}, Vaishaal {Shankar}, Sayna {Ebrahimi}, Trevor {Darrell}, and
  Ludwig {Schmidt}.
\newblock {Predicting with Confidence on Unseen Distributions}.
\newblock In \emph{International Conference on Computer Vision}, 2021.

\bibitem[{Gulrajani} \& {Lopez-Paz}(2020){Gulrajani} and
  {Lopez-Paz}]{gulrajani2020insearch}
Ishaan {Gulrajani} and David {Lopez-Paz}.
\newblock {In Search of Lost Domain Generalization}.
\newblock \emph{arXiv e-prints}, art. arXiv:2007.01434, July 2020.

\bibitem[Gupta et~al.(2021)Gupta, Anpalagan, Guan, and Khwaja]{GUPTA2021100057}
Abhishek Gupta, Alagan Anpalagan, Ling Guan, and Ahmed~Shaharyar Khwaja.
\newblock Deep learning for object detection and scene perception in
  self-driving cars: Survey, challenges, and open issues.
\newblock \emph{Array}, 10:\penalty0 100057, 2021.
\newblock ISSN 2590-0056.
\newblock \doi{https://doi.org/10.1016/j.array.2021.100057}.
\newblock URL
  \url{https://www.sciencedirect.com/science/article/pii/S2590005621000059}.

\bibitem[Haavelmo(1943)]{10.2307/1905714}
Trygve Haavelmo.
\newblock The statistical implications of a system of simultaneous equations.
\newblock \emph{Econometrica}, 11\penalty0 (1):\penalty0 1--12, 1943.
\newblock ISSN 00129682, 14680262.
\newblock URL \url{http://www.jstor.org/stable/1905714}.

\bibitem[{HaoChen} et~al.(2021){HaoChen}, {Wei}, {Gaidon}, and
  {Ma}]{haochen2021provable}
Jeff~Z. {HaoChen}, Colin {Wei}, Adrien {Gaidon}, and Tengyu {Ma}.
\newblock {Provable Guarantees for Self-Supervised Deep Learning with Spectral
  Contrastive Loss}.
\newblock \emph{arXiv e-prints}, art. arXiv:2106.04156, June 2021.

\bibitem[{He} et~al.(2015){He}, {Zhang}, {Ren}, and {Sun}]{he2015resnet}
Kaiming {He}, Xiangyu {Zhang}, Shaoqing {Ren}, and Jian {Sun}.
\newblock {Deep Residual Learning for Image Recognition}.
\newblock \emph{arXiv e-prints}, art. arXiv:1512.03385, December 2015.

\bibitem[Hendrycks et~al.(2020{\natexlab{a}})Hendrycks, Liu, Wallace, Dziedzic,
  Krishnan, and Song]{hendrycks-etal-2020-pretrained}
Dan Hendrycks, Xiaoyuan Liu, Eric Wallace, Adam Dziedzic, Rishabh Krishnan, and
  Dawn Song.
\newblock Pretrained transformers improve out-of-distribution robustness.
\newblock In \emph{Proceedings of the 58th Annual Meeting of the Association
  for Computational Linguistics}, pp.\  2744--2751, Online, July
  2020{\natexlab{a}}. Association for Computational Linguistics.
\newblock \doi{10.18653/v1/2020.acl-main.244}.
\newblock URL \url{https://aclanthology.org/2020.acl-main.244}.

\bibitem[Hendrycks et~al.(2020{\natexlab{b}})Hendrycks, Liu, Wallace, Dziedzic,
  Krishnan, and Song]{hendrycks2020pretrained}
Dan Hendrycks, Xiaoyuan Liu, Eric Wallace, Adam Dziedzic, Rishabh Krishnan, and
  Dawn Song.
\newblock Pretrained transformers improve out-of-distribution robustness.
\newblock In \emph{Association for Computational Linguistics},
  2020{\natexlab{b}}.

\bibitem[Hendrycks et~al.(2021{\natexlab{a}})Hendrycks, Basart, Mu, Kadavath,
  Wang, Dorundo, Desai, Zhu, Parajuli, Guo, Song, Steinhardt, and
  Gilmer]{hendrycks2021many}
Dan Hendrycks, Steven Basart, Norman Mu, Saurav Kadavath, Frank Wang, Evan
  Dorundo, Rahul Desai, Tyler Zhu, Samyak Parajuli, Mike Guo, Dawn Song, Jacob
  Steinhardt, and Justin Gilmer.
\newblock The many faces of robustness: A critical analysis of
  out-of-distribution generalization.
\newblock \emph{ICCV}, 2021{\natexlab{a}}.

\bibitem[Hendrycks et~al.(2021{\natexlab{b}})Hendrycks, Zhao, Basart,
  Steinhardt, and Song]{hendrycks2021nae}
Dan Hendrycks, Kevin Zhao, Steven Basart, Jacob Steinhardt, and Dawn Song.
\newblock Natural adversarial examples.
\newblock \emph{CVPR}, 2021{\natexlab{b}}.

\bibitem[Jiang et~al.(2019)Jiang, Krishnan, Mobahi, and
  Bengio]{jiang2018predicting}
Yiding Jiang, Dilip Krishnan, Hossein Mobahi, and Samy Bengio.
\newblock Predicting the generalization gap in deep networks with margin
  distributions.
\newblock In \emph{International Conference on Learning Representations}, 2019.
\newblock URL \url{https://openreview.net/forum?id=HJlQfnCqKX}.

\bibitem[{Jiang} et~al.(2019){Jiang}, {Neyshabur}, {Mobahi}, {Krishnan}, and
  {Bengio}]{jiang2019fantastic}
Yiding {Jiang}, Behnam {Neyshabur}, Hossein {Mobahi}, Dilip {Krishnan}, and
  Samy {Bengio}.
\newblock {Fantastic Generalization Measures and Where to Find Them}.
\newblock \emph{arXiv e-prints}, art. arXiv:1912.02178, December 2019.

\bibitem[{Jiang} et~al.(2020){Jiang}, {Foret}, {Yak}, {Roy}, {Mobahi},
  {Karolina Dziugaite}, {Bengio}, {Gunasekar}, {Guyon}, and
  {Neyshabur}]{jiang2020predicting}
Yiding {Jiang}, Pierre {Foret}, Scott {Yak}, Daniel~M. {Roy}, Hossein {Mobahi},
  Gintare {Karolina Dziugaite}, Samy {Bengio}, Suriya {Gunasekar}, Isabelle
  {Guyon}, and Behnam {Neyshabur}.
\newblock {NeurIPS 2020 Competition: Predicting Generalization in Deep
  Learning}.
\newblock \emph{arXiv e-prints}, art. arXiv:2012.07976, December 2020.

\bibitem[{Jiang} et~al.(2021){Jiang}, {Nagarajan}, {Baek}, and {Zico
  Kolter}]{2021arXiv210613799J}
Yiding {Jiang}, Vaishnavh {Nagarajan}, Christina {Baek}, and J.~{Zico Kolter}.
\newblock {Assessing Generalization of SGD via Disagreement}.
\newblock \emph{arXiv e-prints}, art. arXiv:2106.13799, June 2021.

\bibitem[Kendall(1938)]{kendall1938new}
M.~G. Kendall.
\newblock A new measure of rank correlation.
\newblock \emph{Biometrika}, 30\penalty0 (1/2):\penalty0 81--93, 1938.
\newblock ISSN 00063444.
\newblock URL \url{http://www.jstor.org/stable/2332226}.

\bibitem[Keskar et~al.(2016)Keskar, Mudigere, Nocedal, Smelyanskiy, and
  Tang]{Keskar2016}
Nitish~Shirish Keskar, Dheevatsa Mudigere, Jorge Nocedal, Mikhail Smelyanskiy,
  and Ping Tak~Peter Tang.
\newblock On large-batch training for deep learning: Generalization gap and
  sharp minima.
\newblock \emph{arXiv preprint arXiv:1609.04836}, 2016.

\bibitem[Kim(2014)]{kim2014convolutional}
Yoon Kim.
\newblock Convolutional neural networks for sentence classification.
\newblock In \emph{Proceedings of the 2014 Conference on Empirical Methods in
  Natural Language Processing (EMNLP)}, 2014.

\bibitem[Kingma \& Ba(2014)Kingma and Ba]{kingma2014method}
Diederik~P. Kingma and Jimmy Ba.
\newblock Adam: A method for stochastic optimization, 2014.
\newblock URL \url{http://arxiv.org/abs/1412.6980}.
\newblock cite arxiv:1412.6980Comment: Published as a conference paper at the
  3rd International Conference for Learning Representations, San Diego, 2015.

\bibitem[Koh et~al.(2021)Koh, Sagawa, Marklund, Xie, Zhang, Balsubramani, Hu,
  Yasunaga, Phillips, Gao, Lee, David, Stavness, Guo, Earnshaw, Haque, Beery,
  Leskovec, Kundaje, Pierson, Levine, Finn, and Liang]{wilds2021}
Pang~Wei Koh, Shiori Sagawa, Henrik Marklund, Sang~Michael Xie, Marvin Zhang,
  Akshay Balsubramani, Weihua Hu, Michihiro Yasunaga, Richard~Lanas Phillips,
  Irena Gao, Tony Lee, Etienne David, Ian Stavness, Wei Guo, Berton~A.
  Earnshaw, Imran~S. Haque, Sara Beery, Jure Leskovec, Anshul Kundaje, Emma
  Pierson, Sergey Levine, Chelsea Finn, and Percy Liang.
\newblock {WILDS}: A benchmark of in-the-wild distribution shifts.
\newblock In \emph{International Conference on Machine Learning (ICML)}, 2021.

\bibitem[Krizhevsky(2009)]{Krizhevsky2009learning}
Alex Krizhevsky.
\newblock Learning multiple layers of features from tiny images.
\newblock Technical report, 2009.

\bibitem[Liang et~al.(2019)Liang, Poggio, Rakhlin, and
  Stokes]{pmlr-v89-liang19a}
Tengyuan Liang, Tomaso Poggio, Alexander Rakhlin, and James Stokes.
\newblock Fisher-rao metric, geometry, and complexity of neural networks.
\newblock In Kamalika Chaudhuri and Masashi Sugiyama (eds.), \emph{Proceedings
  of the Twenty-Second International Conference on Artificial Intelligence and
  Statistics}, volume~89 of \emph{Proceedings of Machine Learning Research},
  pp.\  888--896. PMLR, 16--18 Apr 2019.
\newblock URL \url{https://proceedings.mlr.press/v89/liang19a.html}.

\bibitem[Liang \& Zou(2022)Liang and Zou]{liang2022metashift}
Weixin Liang and James Zou.
\newblock Metashift: A dataset of datasets for evaluating contextual
  distribution shifts and training conflicts.
\newblock In \emph{International Conference on Learning Representations}, 2022.
\newblock URL \url{https://openreview.net/forum?id=MTex8qKavoS}.

\bibitem[{Lin} et~al.(2013){Lin}, {Chen}, and {Yan}]{lin2013network}
Min {Lin}, Qiang {Chen}, and Shuicheng {Yan}.
\newblock {Network In Network}.
\newblock \emph{arXiv e-prints}, art. arXiv:1312.4400, December 2013.

\bibitem[Liu et~al.(2019)Liu, Ott, Goyal, Du, Joshi, Chen, Levy, Lewis,
  Zettlemoyer, and Stoyanov]{liu2019roberta}
Yinhan Liu, Myle Ott, Naman Goyal, Jingfei Du, Mandar Joshi, Danqi Chen, Omer
  Levy, Mike Lewis, Luke Zettlemoyer, and Veselin Stoyanov.
\newblock Roberta: A robustly optimized bert pretraining approach.
\newblock \emph{arXiv preprint arXiv:1907.11692}, 2019.

\bibitem[Lu et~al.(2020)Lu, Nott, Olson, Todeschini, Vahabi, Carmon, and
  Schmidt]{lu2020cifar10.2}
Shangyun Lu, Bradley Nott, Aaron Olson, Alberto Todeschini, Hossein Vahabi,
  Yair Carmon, and Ludwig Schmidt.
\newblock Harder or different? a closer look at distribution shift in dataset
  reproduction.
\newblock In \emph{ICML Workshop on Uncertainty and Robustness in Deep
  Learning}, 2020.

\bibitem[Luo et~al.(2018)Luo, Zhu, Li, Ren, and Zhang]{luo2018smooth}
Yucen Luo, Jun Zhu, Mengxi Li, Yong Ren, and Bo~Zhang.
\newblock Smooth neighbors on teacher graphs for semi-supervised learning.
\newblock In \emph{2018 IEEE/CVF Conference on Computer Vision and Pattern
  Recognition}, pp.\  8896--8905, 2018.
\newblock \doi{10.1109/CVPR.2018.00927}.

\bibitem[McAllester(1999)]{mcallester1999pac}
David~A. McAllester.
\newblock Pac-bayesian model averaging.
\newblock In \emph{Proceedings of the Twelfth Annual Conference on
  Computational Learning Theory}, COLT '99, pp.\  164–170, New York, NY, USA,
  1999. Association for Computing Machinery.
\newblock ISBN 1581131674.
\newblock \doi{10.1145/307400.307435}.
\newblock URL \url{https://doi.org/10.1145/307400.307435}.

\bibitem[Miller et~al.(2021)Miller, Taori, Raghunathan, Sagawa, Koh, Shankar,
  Liang, Carmon, and Schmidt]{miller2021accuracy}
John Miller, Rohan Taori, Aditi Raghunathan, Shiori Sagawa, Pang~Wei Koh,
  Vaishaal Shankar, Percy Liang, Yair Carmon, and Ludwig Schmidt.
\newblock Accuracy on the line: On the strong correlation between
  out-of-distribution and in-distribution generalization, 2021.

\bibitem[Morteza \& Li(2022)Morteza and Li]{morteza2022provable}
Peyman Morteza and Yixuan Li.
\newblock Provable guarantees for understanding out-of-distribution detection.
\newblock \emph{Proceedings of the AAAI Conference on Artificial Intelligence},
  36\penalty0 (7):\penalty0 7831--7840, Jun. 2022.
\newblock \doi{10.1609/aaai.v36i7.20752}.
\newblock URL \url{https://ojs.aaai.org/index.php/AAAI/article/view/20752}.

\bibitem[Mou et~al.(2016)Mou, Men, Li, Xu, Zhang, Yan, and
  Jin]{mou-etal-2016-natural}
Lili Mou, Rui Men, Ge~Li, Yan Xu, Lu~Zhang, Rui Yan, and Zhi Jin.
\newblock Natural language inference by tree-based convolution and heuristic
  matching.
\newblock In \emph{Proceedings of the 54th Annual Meeting of the Association
  for Computational Linguistics (Volume 2: Short Papers)}, pp.\  130--136,
  Berlin, Germany, August 2016. Association for Computational Linguistics.
\newblock \doi{10.18653/v1/P16-2022}.
\newblock URL \url{https://aclanthology.org/P16-2022}.

\bibitem[Nagarajan \& Kolter(2019)Nagarajan and Kolter]{Nagarajan_Kolter_2019}
Vaishnavh Nagarajan and J~Zico Kolter.
\newblock Generalization in deep networks: The role of distance from
  initialization.
\newblock In \emph{NeurIPS Workshop on Deep Learning: Bridging Theory and
  Practice}, 2019.

\bibitem[Nakada \& Imaizumi(2020)Nakada and Imaizumi]{ryumei2020adaptive}
Ryumei Nakada and Masaaki Imaizumi.
\newblock Adaptive approximation and generalization of deep neural network with
  intrinsic dimensionality.
\newblock \emph{Journal of Machine Learning Research}, 21\penalty0
  (174):\penalty0 1--38, 2020.
\newblock URL \url{http://jmlr.org/papers/v21/20-002.html}.

\bibitem[Netzer et~al.(2011)Netzer, Wang, Coates, Bissacco, Wu, and
  Ng]{Netzer2011ReadingDI}
Yuval Netzer, Tao Wang, Adam Coates, A.~Bissacco, Bo~Wu, and A.~Ng.
\newblock Reading digits in natural images with unsupervised feature learning.
\newblock In \emph{NIPS Workshop on Deep Learning and Unsupervised Feature
  Learning}, 2011.

\bibitem[Neyshabur et~al.(2015{\natexlab{a}})Neyshabur, Salakhutdinov, and
  Srebro]{NIPS2015_eaa32c96}
Behnam Neyshabur, Russ~R Salakhutdinov, and Nati Srebro.
\newblock Path-sgd: Path-normalized optimization in deep neural networks.
\newblock In C.~Cortes, N.~Lawrence, D.~Lee, M.~Sugiyama, and R.~Garnett
  (eds.), \emph{Advances in Neural Information Processing Systems}, volume~28.
  Curran Associates, Inc., 2015{\natexlab{a}}.
\newblock URL
  \url{https://proceedings.neurips.cc/paper/2015/file/eaa32c96f620053cf442ad32258076b9-Paper.pdf}.

\bibitem[Neyshabur et~al.(2015{\natexlab{b}})Neyshabur, Tomioka, and
  Srebro]{pmlr-v40-Neyshabur15}
Behnam Neyshabur, Ryota Tomioka, and Nathan Srebro.
\newblock Norm-based capacity control in neural networks.
\newblock In Peter Grünwald, Elad Hazan, and Satyen Kale (eds.),
  \emph{Proceedings of The 28th Conference on Learning Theory}, volume~40 of
  \emph{Proceedings of Machine Learning Research}, pp.\  1376--1401, Paris,
  France, 03--06 Jul 2015{\natexlab{b}}. PMLR.
\newblock URL \url{https://proceedings.mlr.press/v40/Neyshabur15.html}.

\bibitem[Neyshabur et~al.(2017{\natexlab{a}})Neyshabur, Bhojanapalli,
  McAllester, and Srebro]{neyshabur2017exploring}
Behnam Neyshabur, Srinadh Bhojanapalli, David McAllester, and Nathan Srebro.
\newblock Exploring generalization in deep learning.
\newblock In \emph{Proceeding of NeurIPS}, 2017{\natexlab{a}}.

\bibitem[Neyshabur et~al.(2017{\natexlab{b}})Neyshabur, Bhojanapalli, and
  Srebro]{neyshabhur2017pac}
Behnam Neyshabur, Srinadh Bhojanapalli, and Nathan Srebro.
\newblock A pac-bayesian approach to spectrally-normalized margin bounds for
  neural networks.
\newblock In \emph{International Conference on Learning Representations},
  2017{\natexlab{b}}.

\bibitem[Ng et~al.(2020)Ng, Cho, and Ghassemi]{ng2020ssmba}
Nathan Ng, Kyunghyun Cho, and Marzyeh Ghassemi.
\newblock Ssmba: Self-supervised manifold based data augmentation for improving
  out-of-domain robustness.
\newblock In \emph{Proc. of EMNLP}, 2020.

\bibitem[Ni et~al.(2019)Ni, Li, and McAuley]{jianmo2019justifying}
Jianmo Ni, Jiacheng Li, and Julian McAuley.
\newblock Justifying recommendations using distantly-labeled reviews and
  fined-grained aspects.
\newblock In \emph{Proceedings of EMNLP}, 2019.

\bibitem[Ott et~al.(2019)Ott, Edunov, Baevski, Fan, Gross, Ng, Grangier, and
  Auli]{ott2019fairseq}
Myle Ott, Sergey Edunov, Alexei Baevski, Angela Fan, Sam Gross, Nathan Ng,
  David Grangier, and Michael Auli.
\newblock fairseq: A fast, extensible toolkit for sequence modeling.
\newblock In \emph{Proceedings of NAACL-HLT 2019: Demonstrations}, 2019.

\bibitem[Papernot et~al.(2017)Papernot, McDaniel, Goodfellow, Jha, Celik, and
  Swami]{papernot2017practical9}
Nicolas Papernot, Patrick McDaniel, Ian Goodfellow, Somesh Jha, Z.~Berkay
  Celik, and Ananthram Swami.
\newblock Practical black-box attacks against machine learning.
\newblock In \emph{Proceedings of the 2017 ACM on Asia Conference on Computer
  and Communications Security}, ASIA CCS '17, pp.\  506–519, New York, NY,
  USA, 2017. Association for Computing Machinery.
\newblock ISBN 9781450349444.
\newblock \doi{10.1145/3052973.3053009}.
\newblock URL \url{https://doi.org/10.1145/3052973.3053009}.

\bibitem[Pereyra et~al.(2017)Pereyra, Tucker, Chorowski, Kaiser, and
  Hinton]{pereyra2017regularizing}
Gabriel Pereyra, George Tucker, Jan Chorowski, {\L}ukasz Kaiser, and Geoffrey
  Hinton.
\newblock Regularizing neural networks by penalizing confident output
  distributions.
\newblock In \emph{ICLR}, 2017.

\bibitem[Peters et~al.(2015)Peters, Bühlmann, and
  Meinshausen]{peters2015causal}
Jonas Peters, Peter Bühlmann, and Nicolai Meinshausen.
\newblock Causal inference using invariant prediction: identification and
  confidence intervals, 2015.

\bibitem[Radford et~al.(2021)Radford, Kim, Hallacy, Ramesh, Goh, Agarwal,
  Sastry, Askell, Mishkin, Clark, Krueger, and Sutskever]{radford2021clip}
Alec Radford, Jong~Wook Kim, Chris Hallacy, Aditya Ramesh, Gabriel Goh,
  Sandhini Agarwal, Girish Sastry, Amanda Askell, Pamela Mishkin, Jack Clark,
  Gretchen Krueger, and Ilya Sutskever.
\newblock Learning transferable visual models from natural language
  supervision.
\newblock \emph{CoRR}, abs/2103.00020, 2021.
\newblock URL \url{https://arxiv.org/abs/2103.00020}.

\bibitem[Recht et~al.(2018)Recht, Roelofs, Schmidt, and
  Shankar]{recht2018cifar10.1}
Benjamin Recht, Rebecca Roelofs, Ludwig Schmidt, and Vaishaal Shankar.
\newblock Do cifar-10 classifiers generalize to cifar-10?
\newblock 2018.
\newblock \url{https://arxiv.org/abs/1806.00451}.

\bibitem[Recht et~al.(2019)Recht, Roelofs, Schmidt, and
  Shankar]{pmlr-v97-recht19a}
Benjamin Recht, Rebecca Roelofs, Ludwig Schmidt, and Vaishaal Shankar.
\newblock Do {I}mage{N}et classifiers generalize to {I}mage{N}et?
\newblock In Kamalika Chaudhuri and Ruslan Salakhutdinov (eds.),
  \emph{Proceedings of the 36th International Conference on Machine Learning},
  volume~97 of \emph{Proceedings of Machine Learning Research}, pp.\
  5389--5400. PMLR, 09--15 Jun 2019.
\newblock URL \url{https://proceedings.mlr.press/v97/recht19a.html}.

\bibitem[Rico~Sennrich(2016)]{sennrich2016improving}
Alexandra~Birch Rico~Sennrich, Barry~Haddow.
\newblock Improving neural machine translation models with monolingual data.
\newblock In \emph{Proc. of ACL}, 2016.

\bibitem[Rifai et~al.(2011)Rifai, Dauphin, Vincent, Bengio, and
  Muller]{rifai2011manifold}
Salah Rifai, Yann~N Dauphin, Pascal Vincent, Yoshua Bengio, and Xavier Muller.
\newblock The manifold tangent classifier.
\newblock In J.~Shawe-Taylor, R.~Zemel, P.~Bartlett, F.~Pereira, and K.Q.
  Weinberger (eds.), \emph{Advances in Neural Information Processing Systems},
  volume~24. Curran Associates, Inc., 2011.
\newblock URL
  \url{https://proceedings.neurips.cc/paper/2011/file/d1f44e2f09dc172978a4d3151d11d63e-Paper.pdf}.

\bibitem[Romanov \& Shivade(2018)Romanov and Shivade]{romanov2018lessons}
Alexey Romanov and Chaitanya Shivade.
\newblock Lessons from natural language inference in the clinical domain.
\newblock 2018.
\newblock URL \url{http://arxiv.org/abs/1808.06752}.

\bibitem[Russakovsky et~al.(2015)Russakovsky, Deng, Su, Krause, Satheesh, Ma,
  Huang, Karpathy, Khosla, Bernstein, Berg, and Fei-Fei]{russakovsky2015ilsvrc}
Olga Russakovsky, Jia Deng, Hao Su, Jonathan Krause, Sanjeev Satheesh, Sean Ma,
  Zhiheng Huang, Andrej Karpathy, Aditya Khosla, Michael Bernstein,
  Alexander~C. Berg, and Li~Fei-Fei.
\newblock Imagenet large scale visual recognition challenge.
\newblock \emph{International Journal of Computer Vision}, 115\penalty0
  (3):\penalty0 211--252, Dec 2015.
\newblock ISSN 1573-1405.
\newblock \doi{10.1007/s11263-015-0816-y}.
\newblock URL \url{https://doi.org/10.1007/s11263-015-0816-y}.

\bibitem[Sannai et~al.(2021)Sannai, Imaizumi, and Kawano]{sannai2021improved}
Akiyoshi Sannai, Masaaki Imaizumi, and Makoto Kawano.
\newblock Improved generalization bounds of group invariant equivariant deep
  networks via quotient feature spaces.
\newblock In \emph{Conference on Uncertainty in Artificial Intelligence}, 2021.

\bibitem[Schiff et~al.(2021)Schiff, Quanz, Das, and Chen]{schiff2021predicting}
Yair Schiff, Brian Quanz, Payel Das, and Pin-Yu Chen.
\newblock Predicting deep neural network generalization with perturbation
  response curves.
\newblock In \emph{Neural Information Processing Systems}, 2021.

\bibitem[{Schmidt-Hieber}(2019)]{schmidt2019deep}
Johannes {Schmidt-Hieber}.
\newblock {Deep ReLU network approximation of functions on a manifold}.
\newblock \emph{arXiv e-prints}, art. arXiv:1908.00695, August 2019.

\bibitem[Simonyan \& Zisserman(2015)Simonyan and Zisserman]{simonyan2015very}
Karen Simonyan and Andrew Zisserman.
\newblock Very deep convolutional networks for large-scale image recognition.
\newblock In \emph{Proceedings of the International Conference on Learning
  Representations}, 2015.

\bibitem[Sokolić et~al.(2017)Sokolić, Giryes, Sapiro, and Rodrigues]{8024476}
Jure Sokolić, Raja Giryes, Guillermo Sapiro, and Miguel R.~D. Rodrigues.
\newblock Generalization error of deep neural networks: Role of classification
  margin and data structure.
\newblock In \emph{2017 International Conference on Sampling Theory and
  Applications (SampTA)}, pp.\  147--151, 2017.
\newblock \doi{10.1109/SAMPTA.2017.8024476}.

\bibitem[Suzuki(2018)]{suzuki2018adaptivity}
Taiji Suzuki.
\newblock Adaptivity of deep relu network for learning in besov and mixed
  smooth besov spaces: optimal rate and curse of dimensionality.
\newblock In \emph{International Conference on Learning Representations}, 2018.

\bibitem[Suzuki \& Nitanda(2021)Suzuki and Nitanda]{suzuki2021deep}
Taiji Suzuki and Atsushi Nitanda.
\newblock Deep learning is adaptive to intrinsic dimensionality of model
  smoothness in anisotropic besov space.
\newblock In M.~Ranzato, A.~Beygelzimer, Y.~Dauphin, P.S. Liang, and J.~Wortman
  Vaughan (eds.), \emph{Advances in Neural Information Processing Systems},
  volume~34, pp.\  3609--3621. Curran Associates, Inc., 2021.
\newblock URL
  \url{https://proceedings.neurips.cc/paper/2021/file/1dacb10f0623c67cb7dbb37587d8b38a-Paper.pdf}.

\bibitem[Tan \& Le(2019)Tan and Le]{tan2019efficientnet}
Mingxing Tan and Quoc Le.
\newblock {E}fficient{N}et: Rethinking model scaling for convolutional neural
  networks.
\newblock In Kamalika Chaudhuri and Ruslan Salakhutdinov (eds.),
  \emph{Proceedings of the 36th International Conference on Machine Learning},
  volume~97 of \emph{Proceedings of Machine Learning Research}, pp.\
  6105--6114. PMLR, 09--15 Jun 2019.
\newblock URL \url{https://proceedings.mlr.press/v97/tan19a.html}.

\bibitem[Taori et~al.(2020)Taori, Dave, Shankar, Carlini, Recht, and
  Schmidt]{taori2020measuring}
Rohan Taori, Achal Dave, Vaishaal Shankar, Nicholas Carlini, Benjamin Recht,
  and Ludwig Schmidt.
\newblock Measuring robustness to natural distribution shifts in image
  classification.
\newblock In H.~Larochelle, M.~Ranzato, R.~Hadsell, M.F. Balcan, and H.~Lin
  (eds.), \emph{Advances in Neural Information Processing Systems}, volume~33,
  pp.\  18583--18599. Curran Associates, Inc., 2020.
\newblock URL
  \url{https://proceedings.neurips.cc/paper/2020/file/d8330f857a17c53d217014ee776bfd50-Paper.pdf}.

\bibitem[Vapnik \& Chervonenkis(1971)Vapnik and
  Chervonenkis]{vapnik1971uniform}
V.~N. Vapnik and A.~Ya. Chervonenkis.
\newblock On the uniform convergence of relative frequencies of events to their
  probabilities.
\newblock \emph{Theory of Probability \& Its Applications}, 16\penalty0
  (2):\penalty0 264--280, 1971.
\newblock \doi{10.1137/1116025}.
\newblock URL \url{https://doi.org/10.1137/1116025}.

\bibitem[Vedantam et~al.(2021)Vedantam, Lopez-Paz, and
  Schwab]{ramarkishna2021empirical}
Ramakrishna Vedantam, David Lopez-Paz, and David~J. Schwab.
\newblock An empirical investigation of domain generalization with empirical
  risk minimizers.
\newblock In \emph{Neural Information Processing Systems}, 2021.

\bibitem[Verma et~al.(2019)Verma, Lamb, Beckham, Najafi, Mitliagkas, Lopez-Paz,
  and Bengio]{chaudhuri2019manifold}
Vikas Verma, Alex Lamb, Christopher Beckham, Amir Najafi, Ioannis Mitliagkas,
  David Lopez-Paz, and Yoshua Bengio.
\newblock Manifold mixup: Better representations by interpolating hidden
  states.
\newblock In Kamalika Chaudhuri and Ruslan Salakhutdinov (eds.),
  \emph{Proceedings of the 36th International Conference on Machine Learning},
  volume~97 of \emph{Proceedings of Machine Learning Research}, pp.\
  6438--6447. PMLR, 09--15 Jun 2019.
\newblock URL \url{https://proceedings.mlr.press/v97/verma19a.html}.

\bibitem[Wang et~al.(2019)Wang, Ge, Lipton, and Xing]{wang2019imagenetsketch}
Haohan Wang, Songwei Ge, Zachary Lipton, and Eric~P Xing.
\newblock Learning robust global representations by penalizing local predictive
  power.
\newblock In H.~Wallach, H.~Larochelle, A.~Beygelzimer, F.~d\textquotesingle
  Alch\'{e}-Buc, E.~Fox, and R.~Garnett (eds.), \emph{Advances in Neural
  Information Processing Systems}, volume~32. Curran Associates, Inc., 2019.
\newblock URL
  \url{https://proceedings.neurips.cc/paper_files/paper/2019/file/3eefceb8087e964f89c2d59e8a249915-Paper.pdf}.

\bibitem[Wei \& Zou(2019)Wei and Zou]{wei-zou-2019-eda}
Jason Wei and Kai Zou.
\newblock {EDA}: Easy data augmentation techniques for boosting performance on
  text classification tasks.
\newblock In \emph{Proceedings of the 2019 Conference on Empirical Methods in
  Natural Language Processing and the 9th International Joint Conference on
  Natural Language Processing (EMNLP-IJCNLP)}, pp.\  6383--6389, Hong Kong,
  China, November 2019. Association for Computational Linguistics.
\newblock URL \url{https://www.aclweb.org/anthology/D19-1670}.

\bibitem[Wiens et~al.(2019)Wiens, Saria, Sendak, Ghassemi, Liu, Doshi-Velez,
  Jung, Heller, Kale, Saeed, Ossorio, Thadaney-Israni, and Goldenberg]{651011}
J.~Wiens, S.~Saria, M.~Sendak, M.~Ghassemi, V.~Liu, F.~Doshi-Velez, K.~Jung,
  K.~Heller, D.~Kale, M.~Saeed, P.~Ossorio, S.~Thadaney-Israni, and
  A.~Goldenberg.
\newblock Do no harm: A roadmap for responsible machine learning for
  healthcare.
\newblock \emph{Nature Medicine}, 25\penalty0 (10):\penalty0 1337--1340, 2019.

\bibitem[Williams et~al.(2018)Williams, Nangia, and Bowman]{williams2018broad}
Adina Williams, Nikita Nangia, and Samuel Bowman.
\newblock A broad-coverage challenge corpus for sentence understanding through
  inference.
\newblock In \emph{Proceedings of the 2018 Conference of the North American
  Chapter of the Association for Computational Linguistics: Human Language
  Technologies, Volume 1 (Long Papers)}, pp.\  1112--1122. Association for
  Computational Linguistics, 2018.
\newblock URL \url{http://aclweb.org/anthology/N18-1101}.

\bibitem[Wortsman et~al.(2022)Wortsman, Ilharco, Kim, Li, Kornblith, Roelofs,
  Lopes, Hajishirzi, Farhadi, Namkoong, and Schmidt]{wortsman2022robust}
Mitchell Wortsman, Gabriel Ilharco, Jong~Wook Kim, Mike Li, Simon Kornblith,
  Rebecca Roelofs, Raphael~Gontijo Lopes, Hannaneh Hajishirzi, Ali Farhadi,
  Hongseok Namkoong, and Ludwig Schmidt.
\newblock Robust fine-tuning of zero-shot models.
\newblock In \emph{Proceedings of the IEEE/CVF Conference on Computer Vision
  and Pattern Recognition (CVPR)}, pp.\  7959--7971, June 2022.

\bibitem[{Xie} et~al.(2016){Xie}, {Wang}, {Wei}, {Wang}, and
  {Tian}]{2016arXiv160500055X}
Lingxi {Xie}, Jingdong {Wang}, Zhen {Wei}, Meng {Wang}, and Qi~{Tian}.
\newblock {DisturbLabel: Regularizing CNN on the Loss Layer}.
\newblock \emph{arXiv e-prints}, art. arXiv:1605.00055, April 2016.

\bibitem[Xie et~al.(2019)Xie, Dai, Hovy, Luong, and Le]{xie2019unsupervised}
Qizhe Xie, Zihang Dai, Eduard Hovy, Minh-Thang Luong, and Quoc~V Le.
\newblock Unsupervised data augmentation for consistency training.
\newblock \emph{arXiv preprint arXiv:1904.12848}, 2019.

\bibitem[Xie et~al.(2016)Xie, Girshick, Dollár, Tu, and He]{Xie2016resnext}
Saining Xie, Ross Girshick, Piotr Dollár, Zhuowen Tu, and Kaiming He.
\newblock Aggregated residual transformations for deep neural networks.
\newblock \emph{arXiv preprint arXiv:1611.05431}, 2016.

\bibitem[{Yoshida} \& {Miyato}(2017){Yoshida} and
  {Miyato}]{yoshida2017spectral}
Yuichi {Yoshida} and Takeru {Miyato}.
\newblock {Spectral Norm Regularization for Improving the Generalizability of
  Deep Learning}.
\newblock \emph{arXiv e-prints}, art. arXiv:1705.10941, May 2017.

\bibitem[{Zhang} et~al.(2016){Zhang}, {Bengio}, {Hardt}, {Recht}, and
  {Vinyals}]{zhang2016understanding}
Chiyuan {Zhang}, Samy {Bengio}, Moritz {Hardt}, Benjamin {Recht}, and Oriol
  {Vinyals}.
\newblock {Understanding deep learning requires rethinking generalization}.
\newblock \emph{arXiv e-prints}, art. arXiv:1611.03530, November 2016.

\bibitem[Zhang et~al.(2018)Zhang, Cisse, Dauphin, and
  Lopez-Paz]{zhang2018mixup}
Hongyi Zhang, Moustapha Cisse, Yann~N. Dauphin, and David Lopez-Paz.
\newblock mixup: Beyond empirical risk minimization.
\newblock In \emph{International Conference on Learning Representations}, 2018.
\newblock URL \url{https://openreview.net/forum?id=r1Ddp1-Rb}.

\bibitem[Zhang et~al.(2019)Zhang, Liu, Long, and Jordan]{zhang2019bridging}
Yuchen Zhang, Tianle Liu, Mingsheng Long, and Michael Jordan.
\newblock Bridging theory and algorithm for domain adaptation.
\newblock In Kamalika Chaudhuri and Ruslan Salakhutdinov (eds.),
  \emph{Proceedings of the 36th International Conference on Machine Learning},
  volume~97 of \emph{Proceedings of Machine Learning Research}, pp.\
  7404--7413. PMLR, 09--15 Jun 2019.
\newblock URL \url{https://proceedings.mlr.press/v97/zhang19i.html}.

\bibitem[Zhong et~al.(2020)Zhong, Zheng, Kang, Li, and Yang]{zhong2020random}
Zhun Zhong, Liang Zheng, Guoliang Kang, Shaozi Li, and Yi~Yang.
\newblock Random erasing data augmentation.
\newblock \emph{Proceedings of the AAAI Conference on Artificial Intelligence},
  34\penalty0 (07):\penalty0 13001--13008, Apr. 2020.
\newblock \doi{10.1609/aaai.v34i07.7000}.
\newblock URL \url{https://ojs.aaai.org/index.php/AAAI/article/view/7000}.

\bibitem[Zhu et~al.(2021)Zhu, An, and Huang]{zhu2021understanding}
Sicheng Zhu, Bang An, and Furong Huang.
\newblock Understanding the generalization benefit of model invariance from a
  data perspective.
\newblock In \emph{NeurIPS}, 2021.

\end{thebibliography}
\bibliographystyle{tmlr}

\appendix

\section{Experimental Setup}
\label{app:experiment}
In this section we present full details of our experimental setup, including data preprocessing and specifics on model architecture and hyperparameter space.
All models are trained on a single RTX6000 GPU.

\subsection{Data Preprocessing}
\label{app:preprocess}
Large ImageNet scale datasets are preprocessed using the pipeline provided by \citet{taori2020measuring}.
Small scale image classification datasets are preprocessed by normalizing pixel values and resizing to 32 $\times$ 32 if necessary.
We use the same preprocessing steps for sentiment analysis and NLI experiments.
All data is first tokenized using a GPT-2 style tokenizer and BPE vocabulary provided by \texttt{fairseq} \citep{ott2019fairseq}.
This BPE vocabulary consists of 50263 types. 
Corresponding labels are encoded using a label dictionary consisting of as many types as there are classes.
Input text and labels are then binarized for model training. 

\subsection{Model Architecture}
\label{app:app_arch}

The full list of the 196 models we evaluate from the Imagenet-Testbed \citep{taori2020measuring} is provided below:

\scriptsize
\texttt{alexnet\_lpf2, vit\_large\_patch32\_384, wide\_resnet101\_2, resnet50\_aws\_baseline, resnet50\_feature\_cutmix, densenet169, efficientnet-b2-autoaug, vgg11\_bn, resnet50\_with\_jpeg\_compression\_aws, BiT-M-R101x3-nonfinetuned, efficientnet-b6-autoaug, resnet18\_lpf5, mobilenet\_v2, resnet50\_swsl, densenet121\_lpf3, BiT-M-R101x1-nonfinetuned, efficientnet-b2-advprop-autoaug, resnext101\_32x8d\_swsl, resnet50\_with\_fog\_aws, resnet50\_trained\_on\_SIN\_and\_IN, resnext101\_32x4d, resnet50\_with\_contrast\_aws, FixResNet50CutMix, resnet50\_imagenet\_subsample\_1\_of\_32\_batch64\_original\_images, resnext101\_32x4d\_swsl, squeezenet1\_1, resnet50\_imagenet\_subsample\_1\_of\_16\_batch64\_original\_images, resnet18-rotation-nocrop\_40, resnet101\_cutmix, efficientnet-b3, resnet50\_with\_motion\_blur\_aws, vit\_large\_patch16\_384, efficientnet-b4, resnet50\_lpf3, dpn107, resnext101\_32x8d\_deepaugment\_augmix, vgg19, resnet18\_ssl, vgg13\_bn, vgg13, resnet50\_with\_pixelate\_aws, senet154, resnet18\_lpf2, shufflenet\_v2\_x1\_0, se\_resnet101, alexnet\_lpf5, densenet121, efficientnet-b3-advprop-autoaug, resnet50\_augmix, resnet50\_simsiam, efficientnet-b0-advprop-autoaug, resnet50\_imagenet\_subsample\_500\_classes\_batch64\_original\_images, vgg16\_lpf2, mnasnet1\_0, resnet34\_lpf2, dpn68b, mobilenet\_v2\_lpf3, resnet101\_lpf3, alexnet, vgg16\_bn, efficientnet-b0, inceptionv3, resnet18-rotation-worst10\_30, resnet152\_3x\_simclrv2\_finetuned\_100pct\_tf\_port, resnet50\_imagenet\_subsample\_1\_of\_2\_batch64\_original\_images, wide\_resnet50\_2, polynet, efficientnet-b7-randaug, dpn131, vgg16\_bn\_lpf2, instagram-resnext101\_32x16d, vgg16\_bn\_lpf5, resnet50\_linf\_eps8\_robust, efficientnet-b1-advprop-autoaug, inceptionv4, vit\_b\_32\_clip\_zeroshot, resnet18-rotation-worst10\_40, resnet50\_imagenet\_100percent\_batch64\_original\_images, resnet50\_with\_frost\_aws, efficientnet-b3-autoaug, resnet50\_imagenet\_subsample\_125\_classes\_batch64\_original\_images, efficientnet-b7-autoaug, resnet50\_ssl, vgg16\_lpf5, vit\_base\_patch16\_224, resnet34\_lpf5, resnet152, resnext50\_32x4d, FixPNASNet, resnet50\_with\_saturate\_aws, FixResNet50CutMix\_v2, densenet121\_lpf5, resnet50\_imagenet\_subsample\_1\_of\_4\_batch64\_original\_images, resnet18-rotation-random\_40, resnet50\_adv-train-free, resnet18\_lpf3, BiT-M-R50x3-nonfinetuned, efficientnet-b7-advprop-autoaug, resnet50\_with\_spatter\_aws, resnet50\_trained\_on\_SIN, resnet50\_simclrv2\_finetuned\_100pct\_tf\_port, pnasnet5large, BiT-M-R50x3-ILSVRC2012, resnet50\_imagenet\_subsample\_250\_classes\_batch64\_original\_images, efficientnet-b5, resnet50\_deepaugment, efficientnet-b5-randaug, resnet50\_lpf2, se\_resnext50\_32x4d, resnet50\_clip\_zeroshot, resnext50\_32x4d\_swsl, BiT-M-R50x1-nonfinetuned, BiT-M-R101x3-ILSVRC2012, resnet50\_imagenet\_subsample\_1\_of\_8\_batch64\_original\_images, vit\_large\_patch16\_224, efficientnet-b1-autoaug, efficientnet-b6-advprop-autoaug, efficientnet-b5-autoaug, resnet50\_with\_zoom\_blur\_aws, resnext50\_32x4d\_ssl, FixResNet50\_v2, resnet50\_lpf5, resnet101, resnet18\_swsl, efficientnet-b2, squeezenet1\_0, resnet152-imagenet11k, resnet50\_simclrv2\_linear\_probe\_tf\_port, alexnet\_lpf3, bninception, efficientnet-b8-advprop-autoaug, resnet50\_linf\_eps4\_robust, FixResNet50, mnasnet0\_5, resnet50\_mixup, densenet121\_lpf2, resnet18-rotation-standard\_40, se\_resnext101\_32x4d, resnet18-rotation-random\_30, efficientnet-b0-autoaug, efficientnet-b4-autoaug, vgg11, resnext101\_32x8d, BiT-M-R50x1-ILSVRC2012, resnet50, resnet50\_with\_gaussian\_noise\_contrast\_motion\_blur\_jpeg\_compression\_aws, shufflenet\_v2\_x0\_5, dpn92, xception, resnet152\_3x\_simclrv2\_linear\_probe\_tf\_port, dpn98, bninception-imagenet21k, efficientnet-b5-advprop-autoaug, resnext101\_32x16d\_ssl, vit\_base\_patch32\_384, densenet201, inceptionresnetv2, cafferesnet101, instagram-resnext101\_32x8d, resnet34, FixResNet50\_no\_adaptation, resnext101\_32x8d\_ssl, resnet101\_lpf5, mobilenet\_v2\_lpf5, instagram-resnext101\_32x32d, nasnetamobile, mobilenet\_v2\_lpf2, resnet101\_lpf2, se\_resnet50, dpn68, resnet50\_with\_brightness\_aws, resnext101\_64x4d, resnext101\_32x4d\_ssl, vgg19\_bn, fbresnet152, resnet50\_deepaugment\_augmix, se\_resnet152, resnet50\_cutout, resnet50\_cutmix, resnet50\_l2\_eps3\_robust, efficientnet-b1, resnet50\_with\_defocus\_blur\_aws, BiT-M-R101x1-ILSVRC2012, vgg16\_bn\_lpf3, resnet50\_trained\_on\_SIN\_and\_IN\_then\_finetuned\_on\_IN, nasnetalarge, resnet50\_with\_gaussian\_noise\_aws, vit\_base\_patch16\_384, resnet50\_swav, resnet50\_with\_greyscale\_aws, vgg16, resnet34\_lpf3, efficientnet-b4-advprop-autoaug, vgg16\_lpf3, resnet18, densenet161}

\normalsize
Our small image classification models are Network in Network (NiN) \citep{lin2013network}, VGG \citep{simonyan2015very}, and CNN models. 
Training and hyperparemeter details for these models are provided in \citet{jiang2020predicting}.

For natural language tasks,
our CNN models are based on the architecture in \cite{kim2014convolutional}. 
Our input embeddings are 512 dimensional, which we treat as our channel dimension.
Our base model applies a set of three one dimensional convolutions of kernel size 3, 4, and 5 with 256 output channels. 
We modulate the number of stacked convolutions (depth) as well as the channel size (width).
Each convolution generates a separate representation that is max pooled across the sequence and concatenated together.
We feed this representation into a MLP classifier with a single hidden layer of 512 dimensions.
We apply dropout of 0.2 to our inputs and MLP classifier.

Our RoBERTa models use a pre-trained RoBERTa\textsubscript{BASE} model provided by \texttt{fairseq}. 
Classification token embeddings are fed into an MLP classifier with a single hidden layer of 512 dimensions.
All models are written within the \texttt{fairseq} framework \citep{ott2019fairseq} and trained on a single RTX6000 or T4 GPU.

\subsection{Model Hyperparameters}
\label{app:model_hyperparams}

Hyperparameter values for image classification models are provided in \citet{jiang2020predicting}.
For natural language models we vary the following hyperparameters: 
training domain, batch size, depth, width, dropout, weight decay, and label noise.
For training domains, on sentiment analysis we choose between \texttt{books, clothing, home, kindle, movies, pets, sports, tech, tools, toys}. For training domains on NLI, we choose between \texttt{slate, government, fiction, telephone, travel}. NLI datasets include additional test sets \texttt{oup, nineeleven, facetoface, verbatim, letters}.
Possible values for all other hyperparameters are provided in Table \ref{tab:app_hyperparams}

\begin{table*}[t]
\small
\centering
\resizebox{\columnwidth}{!}{%
\begin{tabular}{lcccc}
\toprule
& \multicolumn{2}{c}{CNN} & \multicolumn{2}{c}{RoBERTa} \\
\cmidrule(lr){2-3}
\cmidrule(lr){4-5}
Hyperparameter & SA & NLI & SA & NLI\\
\midrule
Batch Size & \{32, 64, 128\} & \{32, 64, 128 \} & \{8, 16, 32\} & \{8, 16, 32\}\\
Depth & \{1, 2, 3 \} & \{1, 2, 3 \} & 1 & 1 \\
Width & \{128, 256, 512\} & \{128, 256, 512 \} & 768 & 768 \\
Dropout & \{0.0, 0.25, 0.5\} & \{0.0, 0.25\} & \{0.0, 0.1\} & \{0.0, 0.1\}\\
Weight Decay & \{0.0, 0.0001, 0.0005\} & 0.0 & \{0.0, 0.0001, 0.0005\} & \{0.0, 0.0001, 0.0005\}\\
Label Noise & 0.0 & \{0.0, 0.2, 0.4\} & \{0.0, 0.2, 0.4\} & \{0.0, 0.2, 0.4\}\\
\bottomrule
\end{tabular}
}
\caption{Possible hyperparameter values for each architecture and task. }
\label{tab:app_hyperparams}
\end{table*}

\subsection{Model Training}
\label{app:model_train}
All models are trained with the Adam optimizer \citep{kingma2014method} with $\beta = (0.9, 0.98)$ and $\epsilon =$ \num{1e-6}.
CNN models are trained with learning rate \num{1e-3} and RoBERTa models are trained with learning rate \num{1e-5}.
We use a inverse square root learning rate scheduler to anneal learning rate over training.
We early stop CNN models on sentiment analysis at 0.04 cross entropy and on NLI at 0.03 cross entropy. We early stop RoBERTa models on sentiment analysis at 0.05 cross entropy and on NLI at 0.03 cross entropy.
Training details for image classification models are provided in \citet{jiang2020predicting}.

\subsection{Transformation Magnitudes}
\label{sec:app_neighborhood_hyperparams}
We define how to determine the magnitude of each data transformation below, and use a maximum value based on best practices provided in their respective papers.
\begin{itemize}
    \item \textbf{RandAugment: } The magnitude of a transformation is determined by the magnitude parameter in the RandAugment algorithm as well as the number of augmentations applied. In our experiments we consider transformations with a maximum magnitude of 15, with 3 augmentations for larger ImageNet models and 1 augmentation for smaller models.
    \item \textbf{Translate: } The magnitude of a transformation is determined by the maximum percentage of the image the image will be translated in both the X- and Y-axes. In our experiments we consider translations of up to 10\% of the size of the image in both axes.
    \item \textbf{Erase: } The magnitude of the erase transformation is determined by the percentage of the image erased. In our experiments we consider transformations that remove a maximum size of 33\% of the total image area and an aspect ratio between 1/3 and 10/3. 
    \item \textbf{Flip and Crop: } The magnitude of the flip and crop transformation is determined by the flip probability and the crop size. In our experiments we consider transformations that flip the image 50\% of the time and crop the image with a lower bound of 8\% of total image area and an aspect ratio between 3/4 and 4/3. Images are resized to 32 $\times$ 32 after cropping.
    \item \textbf{SSMBA: } The magnitude of a SSMBA transformation is determined by the percentage of tokens corrupted, where of the tokens selected, 10\% are unmasked, 10\% are randomly replaced, and the remaining 80\% are masked. In our experiments we consider SSMBA transformations with a maximum of 15\% of tokens corrupted.
    \item \textbf{EDA: } The magnitude of an EDA transformation is determined by the percentage of tokens noised. In our experiments we consider transformations with a maximum of 10\% of the tokens.
    \item \textbf{Backtranslation: } The magnitude of a backtranslation operation is determined by the temperature of the softmax-ed distribution from which we sample tokens. In our experiments we consider transformations with a maximum temperature of 0.7.
    \item \textbf{Random Replacement: } The magnitude of a random replacement operation is determined by the percentage of tokens replaced. In our experiments we consider transformations with a maximum of 15\% of tokens replaced.
\end{itemize}

\section{Additional Experiments}
\label{app:addl_experiments}

\subsection{Norm-Based Complexity Measures}
\label{app:baseline_norms}
Following \citep{jiang2019fantastic},
we calculate our spectral norm measure as $\Pi_{i=1}^d ||\mathbf{W}_i||_2^2$ and Frobenius norm measure $\Pi_{i=1}^d ||\mathbf{W}_i||_F^2$.
We do not list results on these measures as the correlations are often negative or 0.

\subsection{Cross-Domain Correlation}
\label{app:cross-domain}


In this set of experiments we measure the correlation between neighborhood invariance and generalization values of a single model trained on a single training domain evaluated across different OOD test domains.
For natural language experiments we average correlations across all CNN and RoBERTa models and training domains and call these the \textbf{CNN $\tau$} and \textbf{Roberta $\tau$}.
Since these results are rank correlations over only 9 values, they are quite noisy.
Results are presented in Table \ref{tab:model_tau}.

Neighborhood invariance performs quite poorly on both models, although they still outperforms ATC baselines.
We hypothesize different regions of the input space may have different optimal levels of smoothness that achieve the lowest generalization error.
Our value of interest is then not the absolute smoothness, but the \textit{relative} smoothness compared to this optimal value.
These values are the same when comparing different models evaluated on the same domain, but are not the same for the same model evaluated on different domains, making correlating across domains difficult.
On the natural image manifold where domains are more well behaved and uniform compared to the natural language manifold, the relative smoothness may not differ much between domains allowing us to correlate our measure across domains.

\begin{table}
\small
\centering
\begin{tabular}{llcc}
\toprule
Task & Measure & CNN $\tau$ & RoBERTa $\tau$\\
\midrule
\multirow{6}{*}{SA}
& NI-SSMBA & 0.360 & 0.010\\
& NI-EDA & 0.431 & \textbf{0.266} \\
& NI-BT & 0.505 & 0.245 \\
& NI-RandRep & \textbf{0.570} & 0.150 \\
\cmidrule{2-4}
& ATC-NE & 0.543 & 0.228 \\
& ATC-MC & 0.539 & 0.224 \\
\midrule
\multirow{6}{*}{NLI}
& NI-SSMBA & 0.022 & 0.260 \\
& NI-EDA & 0.102 & 0.335  \\
& NI-BT & 0.089 & 0.333  \\
& NI-RandRep & \textbf{0.226} & \textbf{0.440}\\
\cmidrule{2-4}
& ATC-NE & 0.219 & 0.231 \\
& ATC-MC & 0.223 & 0.239 \\
\bottomrule
\end{tabular}

\caption{Correlation metrics evaluating the ability of our smoothness measure to predict OOD generalization across test datasets. Our smoothness measures achieves strong correlation in image classification tasks but fails in natural language tasks.}
\label{tab:model_tau}
\end{table}

\subsection{Negative Entropy Results}
As an alternative to defining the invariance as the maximum value of the neighborhood decision distribution in Eq. \ref{eq:smoothness}, we also consider defining it using the negative entropy of the same distribution:
\[ \mu(f, x) = \sum_{j\in\Y} p_j(x) \log p_j(x) \]
A full table of results including metrics calculated on neighborhood invariance measured with negative entropy is provided in Table \ref{tab:app_sa_corr_ne}.
We refer to measures calculated with entropy as \textbf{NE-SSMBA}, \textbf{NE-EDA}, \textbf{NE-BT}, and \textbf{NE-Random}. 
For most metrics, NE-* methods perform similarly or slightly worse.

\begin{table*}[h]
\small
\centering
\begin{tabular}{llcccccccc}
\toprule
 & & \multicolumn{4}{c}{CNN} & \multicolumn{4}{c}{RoBERTa}\\
 \cmidrule(lr){3-6}
 \cmidrule(lr){7-10}
Task & Measure & $R^2$ & MAE & Macro $\tau$ & Micro $\tau$ & $R^2$ & MAE & Macro $\tau$ & Micro $\tau$\\
\midrule
\multirow{14}{*}{SA}
& NI-SSMBA & \textbf{0.662} & 1.93 & \textbf{0.677} & \textbf{0.689} & 
\textbf{0.972} & 1.29 & \textbf{0.832} & \textbf{0.829} \\
& NI-EDA & 0.641 & 2.04 & 0.664  & 0.649 & 
0.968 & 1.45 & 0.830  & 0.810 \\
& NI-BT & 0.550 & 2.99 & 0.592  & 0.501 & 
0.961 & 1.47 & 0.813  & 0.801 \\
& NI-RandRep & 0.409 & 2.64 & 0.544 & 0.554 & 
0.967 & \textbf{1.27} & 0.821 & 0.816 \\
\cmidrule{2-10}
& NE-SSMBA & 0.595 & 2.53 & 0.708 & 0.713 & 0.971 & 1.32 & 0.830 & 0.824 \\
& NE-EDA & 0.534 & 2.71 & 0.698 & 0.674 & 0.965 & 1.55 & 0.825 & 0.801 \\
& NE-BT & 0.471 & 3.59 & 0.618 & 0.541 & 0.961 & 1.46 & 0.813 & 0.799 \\
& NE-RandRep & 0.283 & 3.37 & 0.570 & 0.552 & 0.964 & 1.34 & 0.818 & 0.809 \\
\cmidrule{2-10}
& ATC-NE & 0.530 & 3.80 & 0.506 & 0.642 &
0.849 & 3.59 & 0.684 & 0.706 \\
& ATC-MC & 0.528 & 3.76 & 0.507 & 0.642 & 
0.863 & 3.54 & 0.698 & 0.716 \\
\midrule
\multirow{14}{*}{NLI}
& NI-SSMBA & 0.575 & 2.09 & 0.570 & \textbf{0.534} &
0.933 & 1.19 & 0.750 & 0.730 \\
& NI-EDA & 0.577 & \textbf{2.04} & 0.581 & 0.511 &
0.941 & 1.26 & \textbf{0.789} & \textbf{0.757} \\
& NI-BT & 0.509 & 2.11 & 0.470 & 0.449 &
0.944 & 1.07 & 0.759 & 0.740 \\
& NI-RandRep & 0.451 & 2.20 & 0.452 & 0.428 &
0.890 & 1.70 & 0.688 & 0.647\\
\cmidrule{2-10}
& NE-SSMBA & 0.588 & 2.20 & 0.579 & 0.520 & 0.941 & 1.39 & 0.738 & 0.711 \\
& NE-EDA & \textbf{0.606} & 2.11 & \textbf{0.597} & 0.512 & 0.937 & 1.48 & 0.767 & 0.732 \\
& NE-BT & 0.536 & 2.27 & 0.480 & 0.422 & \textbf{0.954} & \textbf{1.12} & 0.764 & 0.750 \\
& NE-RandRep & 0.457 & 2.36 & 0.451 & 0.397 & 0.904 & 1.79 & 0.665 & 0.591 \\
\cmidrule{2-10}
& ATC-NE & 0.378 & 3.57 & 0.430 & 0.294 &
0.673 & 2.35 & 0.536 & 0.52 \\
& ATC-MC & 0.382 & 3.57 & 0.433 & 0.297 &
0.718 & 2.21 & 0.570 & 0.556\\
\bottomrule
\end{tabular}
\caption{Correlation metrics evaluating the quality of our neighborhood invariance measure on two tasks, sentiment analysis and natural language inference, and two architectures, CNN and RoBERTa. Details on metric calculations and baselines are provided in sections \ref{sec:metrics} and \ref{sec:baselines}. This full table of results includes metrics calculated with neighborhood negative entropy measure as well.}
\label{tab:app_sa_corr_ne}
\end{table*}

\section{Full Results}
\label{app:fullres}
We provide a full breakdown of results on the correlation metrics $R^2$ (Tables \ref{tab:app_full_r2_svhn}, \ref{tab:app_full_r2_cifar}, \ref{tab:app_full_r2_cnn_aws}, \ref{tab:app_full_r2_bert_aws}, \ref{tab:app_full_r2_cnn_nli}, \ref{tab:app_full_r2_bert_nli}), macro $\tau$ (Tables \ref{tab:app_full_macrot_svhn}, \ref{tab:app_full_macrot_cifar}, \ref{tab:app_full_macrot_cnn_aws}, \ref{tab:app_full_macrot_bert_aws}, \ref{tab:app_full_macrot_cnn_nli}, \ref{tab:app_full_macrot_bert_nli}, micro $\tau$ (Tables \ref{tab:app_full_microt_aws}, \ref{tab:app_full_microt_nli}), and ID $\tau$ (Tables \ref{tab:app_full_idt_image}, \ref{tab:app_full_idt_aws}, \ref{tab:app_full_idt_nli}) for each set of datasets and models. 
We also provide additional standard deviations for all main results in Table \ref{tab:res_full}.

\begin{table*}
\centering
\scriptsize
\begin{tabular}{lccccccc}
\toprule
& \multicolumn{7}{c}{Test Domain}\\
\cmidrule{1-8}
Measure & ImageNetV2 & ImagenNet-Sketch & ObjectNet & ImageNet-Vid & YTBB & ImageNet-R & ImageNet-A \\
\midrule
NI-RandAug & 0.810 & 0.641 & \textbf{0.613} & \textbf{0.767} & \textbf{0.570} & \textbf{0.763} & 0.577\\
NI-Translate & 0.794 & 0.436 & 0.461 & 0.642 & 0.395 & 0.560 & 0.468\\
NI-Erase & 0.715 & 0.222 & 0.384 & 0.635 & 0.399 & 0.398 & 0.446\\
NI-FC & 0.767 & 0.406 & 0.516 & 0.706 & 0.446 & 0.618 & \textbf{0.679}\\
\midrule
ATC-MC & 0.980 & \textbf{0.710} & 0.451 & 0.552 & 0.200 & 0.463 & 0.159\\
ATC-NE & \textbf{0.991} & 0.709 & 0.392 & 0.484 & 0.229 & 0.441 & 0.209\\
\bottomrule
\end{tabular}
\caption{Full $R^2$ metrics for all ImageNet test domains. We average values across models.}
\label{tab:app_full_r2_imagenet}
\end{table*}

\begin{table*}[h]
\centering
\scriptsize
\begin{tabular}{lllccc}
\toprule
& & &\multicolumn{3}{c}{Test Domain}\\
\cmidrule{4-6}
Model & Train Domain & Measure & SVHN & Colored MNIST & MNIST\\
\midrule
\multirow{5}{*}{NiN} & \multirow{5}{*}{SVHN}
& NI-RandAug  & \textemdash  & \textbf{0.785} & 0.744\\
& &  NI-Translate  & \textemdash  & 0.728 & 0.642 \\
& & NI-Erase  & \textemdash  & 0.088 & 0.218\\
& & NI-FC  & \textemdash  & 0.134 & 0.283\\
\cmidrule{3-6}
& & ATC-NE  & \textemdash  & 0.506 & 0.725\\
& & ATC-MC  & \textemdash  & 0.615 & \textbf{0.769}\\
\bottomrule
\end{tabular}
\caption{Full $R^2$ metrics for all test domains for image classification models on SVHN, Colored MNIST, and MNIST.}
\label{tab:app_full_r2_svhn}
\end{table*}

\begin{table*}[h]
\centering
\scriptsize
\begin{tabular}{lllcccc}
\toprule
& & &\multicolumn{4}{c}{Test Domain}\\
\cmidrule{4-7}
Model & Train Domain & Measure & CIFAR10 & CINIC10 & CIFAR10.1 & CIFAR10.2\\
\midrule
\multirow{5}{*}{NiN} & \multirow{5}{*}{CIFAR10}
& NI-RandAug  & \textemdash  & \textbf{0.898} & \textbf{0.927} & \textbf{0.876}\\
& & NI-Translate  & \textemdash  & 0.688 & 0.864 & 0.730 \\
& & NI-Erase  & \textemdash  & 0.404 & 0.526 & 0.547\\
& & NI-FC  & \textemdash  & 0.109 & 0.028 & 0.000\\
\cmidrule{3-7}
& & ATC-NE  & \textemdash  & 0.237 & 0.575 & 0.435\\
& & ATC-MC  & \textemdash  & 0.252 & 0.538 & 0.390\\
\midrule
\multirow{5}{*}{ResNet} & \multirow{5}{*}{CIFAR10}
& NI-RandAug  & \textemdash  & 0.816 & 0.844 & \textbf{0.853} \\
& & NI-Translate & \textemdash  & \textbf{0.889} & \textbf{0.899} & 0.836\\
& & NI-Erase  & \textemdash  & 0.807 & 0.812 & 0.783 \\
& & NI-FC  & \textemdash  & 0.726 & 0.628 & 0.660 \\
\cmidrule{3-7}
& & ATC-NE  & \textemdash & 0.736 & 0.782 & 0.693 \\
& & ATC-MC  & \textemdash & 0.702 & 0.760 & 0.680 \\
\midrule
\multirow{5}{*}{VGG} & \multirow{5}{*}{CIFAR10}
& NI-RandAug  & \textemdash  & \textbf{0.969} & \textbf{0.950} & \textbf{0.929}\\
& & NI-Translate  & \textemdash  & 0.849 & 0.844 & 0.809 \\
& & NI-Erase  & \textemdash  & 0.096 & 0.100 & 0.104\\
& & NI-FC  & \textemdash  & 0.637 & 0.524 & 0.487\\
\cmidrule{3-7}
& & ATC-NE  & \textemdash  & 0.557 & 0.774 & 0.724\\
& & ATC-MC  & \textemdash  & 0.559 & 0.764 & 0.709\\
\midrule
\multirow{5}{*}{CNN} & \multirow{5}{*}{CINIC10}
& NI-RandAug  & \textbf{0.922} & \textemdash  & \textbf{0.876} & \textbf{0.865}\\
& & NI-Translate  & 0.516 & \textemdash  & 0.449 & 0.407 \\
& & NI-Erase & 0.603 & \textemdash  & 0.504 & 0.548 \\
& & NI-FC  & 0.416 & \textemdash  & 0.397 & 0.395\\
\cmidrule{3-7}
& & ATC-NE  & 0.869 & \textemdash  & 0.750 & 0.724\\
& & ATC-MC  & 0.868 & \textemdash  & 0.741 & 0.716\\
\bottomrule
\end{tabular}
\caption{Full $R^2$ metrics for all test domains for image classification models on CIFAR10, CINIC10, CIFAR10.1, and CIFAR10.2.}
\label{tab:app_full_r2_cifar}
\end{table*}

\begin{table*}
\centering
\resizebox*{!}{0.9\textheight}{%
\begin{tabular}{llcccccccccc}
\toprule
& & \multicolumn{10}{c}{Test Domain}\\
\cmidrule{3-12}
Train Domain & Measure & \texttt{books} & \texttt{clothing} & \texttt{home} & \texttt{kindle} & \texttt{movies} & \texttt{pets} & \texttt{sports} & \texttt{tech} & \texttt{tools} & \texttt{toys} \\
\midrule
\multirow{6}{*}{\texttt{books}} 
& NI-SSMBA & \textemdash & 0.688 & 0.771 & 0.752 & 0.683 & 0.823 & 0.762 & 0.697 & 0.711 & 0.729\\
& NI-EDA & \textemdash & 0.720 & 0.725 & 0.699 & 0.689 & 0.810 & 0.693 & 0.663 & 0.727 & 0.805\\
& NI-BT & \textemdash & 0.560 & 0.663 & 0.639 & 0.593 & 0.675 & 0.602 & 0.591 & 0.592 & 0.658\\
& NI-RandRep & \textemdash & 0.413 & 0.418 & 0.331 & 0.343 & 0.479 & 0.373 & 0.304 & 0.401 & 0.496\\
\cmidrule{2-12}
& ATC-NE & \textemdash & \textbf{0.765} & \textbf{0.859} & \textbf{0.828} & \textbf{0.777} & \textbf{0.834} & \textbf{0.834} & \textbf{0.791} & \textbf{0.814} & \textbf{0.872}\\
& ATC-MC & \textemdash & 0.759 & 0.854 & 0.821 & 0.771 & 0.828 & 0.830 & 0.790 & 0.808 & 0.871\\
\bottomrule
\end{tabular}
}
\caption{Full $R^2$ metrics for all pairs of training and test domains for CNN models trained on AWS.}
\label{tab:app_full_r2_cnn_aws}
\end{table*}

\begin{table*}
\centering
\resizebox*{!}{0.9\textheight}{%
\begin{tabular}{llcccccccccc}
\toprule
& & \multicolumn{10}{c}{Test Domain}\\
\cmidrule{3-12}
Train Domain & Measure & \texttt{books} & \texttt{clothing} & \texttt{home} & \texttt{kindle} & \texttt{movies} & \texttt{pets} & \texttt{sports} & \texttt{tech} & \texttt{tools} & \texttt{toys} \\
\midrule
\multirow{6}{*}{\texttt{books}} 
&  NI-SSMBA  & \textemdash & \textbf{0.973} & \textbf{0.983} & \textbf{0.987} & 0.963 & \textbf{0.973} & \textbf{0.985} & \textbf{0.985} & 0.977 & 0.968\\
&  NI-EDA  & \textemdash & 0.961 & 0.970 & 0.982 & \textbf{0.966} & 0.966 & 0.970 & 0.957 & \textbf{0.978} & \textbf{0.981}\\
&  NI-BT  & \textemdash & 0.946 & 0.979 & 0.985 & 0.961 & 0.972 & 0.972 & 0.975 & 0.970 & 0.955\\
&  NI-RandRep  & \textemdash & 0.956 & 0.970 & 0.981 & 0.958 & 0.967 & 0.970 & 0.973 & 0.952 & 0.961\\
\cmidrule{2-12}
&  ATC-NE  & \textemdash & 0.845 & 0.861 & 0.957 & 0.927 & 0.940 & 0.852 & 0.925 & 0.853 & 0.876\\
&  ATC-MC  & \textemdash & 0.838 & 0.842 & 0.947 & 0.924 & 0.929 & 0.840 & 0.927 & 0.845 & 0.876\\
\bottomrule
\end{tabular}
}
\caption{Full $R^2$ metrics for all pairs of training and test domains for BERT models trained on AWS.}
\label{tab:app_full_r2_bert_aws}
\end{table*}

\begin{table*}
\resizebox{1\columnwidth}{!}{%
\begin{tabular}{llcccccccccc}
\toprule
& & \multicolumn{10}{c}{Test Domain}\\
\cmidrule{3-12}
Train Domain & Measure & \texttt{slate} & \texttt{verbatim} & \texttt{facetoface} & \texttt{oup} & \texttt{nineeleven} & \texttt{fiction} & \texttt{telephone} & \texttt{travel} & \texttt{letters} & \texttt{government} \\
\midrule
\multirow{6}{*}{\texttt{slate}} 
&  NI-SSMBA  & \textemdash & 0.457 & 0.760 & 0.544 & 0.642 & 0.706 & 0.727 & 0.585 & 0.640 & 0.714\\
&  NI-EDA  & \textemdash & 0.574 & 0.723 & 0.565 & 0.620 & 0.708 & 0.731 & 0.549 & 0.595 & 0.706\\
&  NI-BT  & \textemdash & \textbf{0.862} & \textbf{0.920} & \textbf{0.942} & \textbf{0.930} & \textbf{0.913} & \textbf{0.878} & \textbf{0.936} & \textbf{0.949} & \textbf{0.940}\\
&  NI-RandRep  & \textemdash & 0.242 & 0.336 & 0.501 & 0.415 & 0.121 & 0.170 & 0.301 & 0.652 & 0.470\\
\cmidrule{2-12}
&  ATC-NE  & \textemdash & 0.681 & 0.677 & 0.594 & 0.599 & 0.741 & 0.669 & 0.664 & 0.661 & 0.773\\
&  ATC-MC  & \textemdash & 0.682 & 0.675 & 0.590 & 0.600 & 0.737 & 0.670 & 0.665 & 0.661 & 0.771\\
\bottomrule
\end{tabular}
}
\caption{Full $R^2$ metrics for all pairs of training and test domains for CNN models trained on MNLI.}
\label{tab:app_full_r2_cnn_nli}
\end{table*}

\begin{table*}
\resizebox{1\columnwidth}{!}{%
\begin{tabular}{llcccccccccc}
\toprule
& & \multicolumn{10}{c}{Test Domain}\\
\cmidrule{3-12}
Train Domain & Measure & \texttt{slate} & \texttt{verbatim} & \texttt{facetoface} & \texttt{oup} & \texttt{nineeleven} & \texttt{fiction} & \texttt{telephone} & \texttt{travel} & \texttt{letters} & \texttt{government} \\
\midrule
\multirow{6}{*}{\texttt{slate}} 
&  NI-SSMBA  & \textemdash & \textbf{0.920} & 0.905 & 0.943 & 0.916 & 0.929 & \textbf{0.930} & \textbf{0.957} & 0.931 & \textbf{0.970}\\
&  NI-EDA  & \textemdash & 0.701 & 0.754 & 0.816 & 0.784 & 0.790 & 0.708 & 0.787 & 0.820 & 0.837\\
&  NI-BT  & \textemdash & 0.862 & \textbf{0.920} & 0.942 & \textbf{0.930} & 0.913 & 0.878 & 0.936 & \textbf{0.949} & 0.940\\
&  NI-RandRep  & \textemdash & 0.242 & 0.336 & 0.501 & 0.415 & 0.121 & 0.170 & 0.301 & 0.652 & 0.470\\
\cmidrule{2-12}
&  ATC-NE  & \textemdash & 0.878 & 0.899 & \textbf{0.954} & 0.911 & 0.925 & 0.911 & 0.904 & 0.890 & 0.951\\
&  ATC-MC  & \textemdash & 0.871 & 0.884 & 0.953 & 0.908 & \textbf{0.932} & 0.914 & 0.911 & 0.887 & 0.949\\
\bottomrule
\end{tabular}
}
\caption{Full $R^2$ metrics for all pairs of training and test domains for BERT models trained on MNLI.}
\label{tab:app_full_r2_bert_nli}
\end{table*}

\begin{table*}
\centering
\scriptsize
\begin{tabular}{lccccccc}
\toprule
& \multicolumn{7}{c}{Test Domain}\\
\cmidrule{1-8}
Measure & ImageNetV2 & ImagenNet-Sketch & ObjectNet & ImageNet-Vid & YTBB & ImageNet-R & ImageNet-A \\
\midrule
NI-RandAug & 0.816 & 0.677 & 0.646 & \textbf{0.790} & \textbf{0.692} & \textbf{0.724} & 0.586\\
NI-Translate & 0.785 & 0.515 & 0.516 & 0.685 & 0.544 & 0.581 & 0.439\\
NI-Erase & 0.779 & 0.322 & 0.52 & 0.706 & 0.560 & 0.444 & 0.517\\
NI-FC & 0.842 & 0.552 & 0.645 & 0.780 & 0.649 & 0.679 & \textbf{0.589}\\
\midrule
ATC-MC & 0.891 & 0.708 & \textbf{0.756} & 0.761 & 0.509 & 0.521 & 0.190\\
ATC-NE & \textbf{0.926} & \textbf{0.717} & 0.671 & 0.651 & 0.494 & 0.569 & 0.248\\
\bottomrule
\end{tabular}
\caption{Full macro $\tau$ metrics for all ImageNet test domains. We average values across models.}
\label{tab:app_full_macrot_imagenet}
\end{table*}

\begin{table*}
\centering
\scriptsize
\begin{tabular}{lllccc}
\toprule
& & &\multicolumn{3}{c}{Test Domain}\\
\cmidrule{4-6}
Model & Train Domain & Measure & SVHN & Colored MNIST & MNIST\\
\midrule
\multirow{5}{*}{NiN} & \multirow{5}{*}{SVHN}
& NI-RandAug  & \textemdash  & 0.668 & 0.616\\
& & NI-Translate  & \textemdash  & \textbf{0.699} & 0.635 \\
& & NI-Erase & \textemdash  & -0.102 & -0.168\\
& & NI-FC  & \textemdash  & -0.233 & -0.398\\
\cmidrule{3-6}
& & ATC-NE  & \textemdash  & 0.577 & 0.695\\
& & ATC-MC  & \textemdash  & 0.646 & \textbf{0.716}\\
\bottomrule
\end{tabular}
\caption{Full macro $\tau$ metrics for all test domains for image classification models on SVHN, Colored MNIST, and MNIST.}
\label{tab:app_full_macrot_svhn}
\end{table*}

\begin{table*}
\centering
\scriptsize
\begin{tabular}{lllcccc}
\toprule
& & &\multicolumn{4}{c}{Test Domain}\\
\cmidrule{4-7}
Model & Train Domain & Measure & CIFAR10 & CINIC10 & CIFAR10.1 & CIFAR10.2\\
\midrule
\multirow{5}{*}{NiN} & \multirow{5}{*}{CIFAR10}
& NI-RandAug  & \textemdash  & \textbf{0.785} & \textbf{0.830} & \textbf{0.783}\\
& & NI-Translate  & \textemdash  & 0.621 & 0.739 & 0.644 \\
& & NI-Erase  & \textemdash  & 0.385 & 0.485 & 0.499\\
& & NI-FC  & \textemdash  & 0.194 & 0.077 & -0.037\\
\cmidrule{3-7}
& & ATC-NE  & \textemdash  & 0.250 & 0.543 & 0.484\\
& & ATC-MC  & \textemdash  & 0.319 & 0.514 & 0.446\\
\midrule
\multirow{5}{*}{ResNet} & \multirow{5}{*}{CIFAR10}
& NI-RandAug  & \textemdash  & 0.706 & 0.722 & \textbf{0.742}\\
& & NI-Translate    & \textemdash  & \textbf{0.800} & \textbf{0.796} & 0.741 \\
& & NI-Erase  & \textemdash  & 0.723 & 0.727 & 0.699 \\
& & NI-FC  & \textemdash  & 0.647 & 0.585 & 0.607\\
\cmidrule{3-7}
& & ATC-NE  & \textemdash & 0.644 & 0.687 & 0.630 \\
& & ATC-MC  & \textemdash & 0.631 & 0.672 & 0.624 \\
\midrule
\multirow{5}{*}{VGG} & \multirow{5}{*}{CIFAR10}
& NI-RandAug  & \textemdash  & \textbf{0.868} & \textbf{0.831} & \textbf{0.772}\\
& & NI-Translate  & \textemdash  & 0.700 & 0.679 & 0.632 \\
& & NI-Erase  & \textemdash  & -0.153 & -0.196 & -0.215\\
& & NI-FC  & \textemdash  & 0.597 & 0.569 & 0.512\\
\cmidrule{3-7}
& & ATC-NE  & \textemdash  & 0.531 & 0.656 & 0.599\\
& & ATC-MC  & \textemdash  & 0.533 & 0.648 & 0.586\\
\midrule
\multirow{5}{*}{CNN} & \multirow{5}{*}{CINIC10}
& NI-RandAug  & \textbf{0.756} & \textemdash  & \textbf{0.698} & \textbf{0.695}\\
& & NI-Translate  & 0.347 & \textemdash  & 0.318 & 0.267 \\
& & NI-Erase  & 0.619 & \textemdash  & 0.539 & 0.577\\
& & NI-FC  & 0.249 & \textemdash  & 0.286 & 0.252\\
\cmidrule{3-7}
& & ATC-NE  & 0.607 & \textemdash  & 0.510 & 0.447\\
& & ATC-MC  & 0.611 & \textemdash  & 0.500 & 0.440\\
\bottomrule
\end{tabular}
\caption{Full macro $\tau$ metrics for all test domains for image classification models on CIFAR10, CINIC10, CIFAR10.1, and CIFAR10.2.}
\label{tab:app_full_macrot_cifar}
\end{table*}

\begin{table*}
\centering
\resizebox*{!}{0.9\textheight}{%
\begin{tabular}{llcccccccccc}
\toprule
& & \multicolumn{10}{c}{Test Domain}\\
\cmidrule{3-12}
Train Domain & Measure & \texttt{books} & \texttt{clothing} & \texttt{home} & \texttt{kindle} & \texttt{movies} & \texttt{pets} & \texttt{sports} & \texttt{tech} & \texttt{tools} & \texttt{toys} \\
\midrule
\multirow{6}{*}{\texttt{books}} 
&  NI-SSMBA  & \textemdash & 0.708 & \textbf{0.741} & \textbf{0.687} & 0.678 & 0.754 & \textbf{0.738} & \textbf{0.738} & \textbf{0.721} & 0.669\\
&  NI-EDA  & \textemdash & \textbf{0.722} & 0.738 & 0.640 & \textbf{0.685} & \textbf{0.756} & 0.688 & 0.709 & 0.719 & \textbf{0.725}\\
&  NI-BT  & \textemdash & 0.615 & 0.631 & 0.549 & 0.596 & 0.656 & 0.602 & 0.644 & 0.584 & 0.621\\
&  NI-RandRep  & \textemdash & 0.591 & 0.586 & 0.499 & 0.587 & 0.587 & 0.530 & 0.530 & 0.546 & 0.589\\
\cmidrule{2-12}
&  ATC-NE  & \textemdash & 0.605 & 0.642 & 0.534 & 0.496 & 0.629 & 0.605 & 0.538 & 0.618 & 0.673\\
&  ATC-MC  & \textemdash & 0.604 & 0.639 & 0.534 & 0.489 & 0.630 & 0.606 & 0.544 & 0.616 & 0.670\\
\bottomrule
\end{tabular}
}
\caption{Full macro $\tau$ metrics for all pairs of training and test domains for CNN models trained on AWS.}
\label{tab:app_full_macrot_cnn_aws}
\end{table*}

\begin{table*}
\centering
\resizebox*{!}{0.9\textheight}{%
\begin{tabular}{llcccccccccc}
\toprule
& & \multicolumn{10}{c}{Test Domain}\\
\cmidrule{3-12}
Train Domain & Measure & \texttt{books} & \texttt{clothing} & \texttt{home} & \texttt{kindle} & \texttt{movies} & \texttt{pets} & \texttt{sports} & \texttt{tech} & \texttt{tools} & \texttt{toys} \\
\midrule
\multirow{6}{*}{\texttt{books}} 
&  NI-SSMBA  & \textemdash & \textbf{0.886} & \textbf{0.906} & 0.893 & 0.901 & \textbf{0.900} & \textbf{0.925} & \textbf{0.918} & \textbf{0.931} & \textbf{0.857}\\
&  NI-EDA  & \textemdash & 0.883 & 0.851 & 0.890 & 0.852 & 0.863 & 0.854 & 0.855 & 0.895 & 0.844\\
&  NI-BT  & \textemdash & 0.830 & 0.887 & 0.893 & 0.883 & 0.884 & 0.871 & 0.878 & 0.869 & 0.824\\
&  NI-RandRep  & \textemdash & 0.882 & 0.871 & \textbf{0.908} & \textbf{0.906} & 0.897 & 0.894 & 0.900 & 0.913 & 0.846\\
\cmidrule{2-12}
&  ATC-NE  & \textemdash & 0.803 & 0.814 & 0.851 & 0.821 & 0.885 & 0.756 & 0.843 & 0.774 & 0.832\\
&  ATC-MC  & \textemdash & 0.789 & 0.806 & 0.850 & 0.794 & 0.874 & 0.760 & 0.845 & 0.768 & 0.824\\
\bottomrule
\end{tabular}
}
\caption{Full macro $\tau$ metrics for all pairs of training and test domains for BERT models trained on AWS.}
\label{tab:app_full_macrot_bert_aws}
\end{table*}

\begin{table*}
\resizebox{1\columnwidth}{!}{%
\begin{tabular}{llcccccccccc}
\toprule
& & \multicolumn{10}{c}{Test Domain}\\
\cmidrule{3-12}
Train Domain & Measure & \texttt{slate} & \texttt{verbatim} & \texttt{facetoface} & \texttt{oup} & \texttt{nineeleven} & \texttt{fiction} & \texttt{telephone} & \texttt{travel} & \texttt{letters} & \texttt{government} \\
\midrule
\multirow{6}{*}{\texttt{slate}} 
&  NI-SSMBA  & \textemdash & 0.483 & 0.675 & 0.553 & 0.608 & 0.650 & 0.661 & 0.570 & 0.622 & 0.641\\
&  NI-EDA  & \textemdash & 0.564 & 0.675 & 0.563 & 0.599 & 0.658 & 0.672 & 0.557 & 0.599 & 0.649\\
&  NI-BT  & \textemdash & \textbf{0.669} & \textbf{0.742} & \textbf{0.736} & \textbf{0.756} & \textbf{0.727} & \textbf{0.754} & \textbf{0.751} & \textbf{0.738} & \textbf{0.791}\\
&  NI-RandRep  & \textemdash & 0.393 & 0.526 & 0.529 & 0.480 & 0.398 & 0.438 & 0.445 & 0.640 & 0.531\\
\cmidrule{2-12}
&  ATC-NE  & \textemdash & 0.650 & 0.632 & 0.577 & 0.594 & 0.683 & 0.625 & 0.602 & 0.621 & 0.699\\
&  ATC-MC  & \textemdash & 0.652 & 0.632 & 0.578 & 0.594 & 0.681 & 0.625 & 0.603 & 0.622 & 0.697\\
\bottomrule
\end{tabular}
}
\caption{Full macro $\tau$ metrics for all pairs of training and test domains for CNN models trained on MNLI.}
\label{tab:app_full_macrot_cnn_nli}
\end{table*}

\begin{table*}
\resizebox{1\columnwidth}{!}{%
\begin{tabular}{llcccccccccc}
\toprule
& & \multicolumn{10}{c}{Test Domain}\\
\cmidrule{3-12}
Train Domain & Measure & \texttt{slate} & \texttt{verbatim} & \texttt{facetoface} & \texttt{oup} & \texttt{nineeleven} & \texttt{fiction} & \texttt{telephone} & \texttt{travel} & \texttt{letters} & \texttt{government} \\
\midrule
\multirow{6}{*}{\texttt{slate}} 
&  NI-SSMBA  & \textemdash & \textbf{0.743} & 0.746 & 0.709 & \textbf{0.767} & 0.742 & 0.727 & \textbf{0.771} & \textbf{0.782} & \textbf{0.819}\\
&  NI-EDA  & \textemdash & 0.580 & 0.695 & 0.674 & 0.696 & 0.703 & 0.655 & 0.651 & 0.671 & 0.737\\
&  NI-BT  & \textemdash & 0.669 & 0.742 & 0.736 & 0.756 & 0.727 & 0.754 & 0.751 & 0.738 & 0.791\\
&  NI-RandRep  & \textemdash & 0.393 & 0.526 & 0.529 & 0.480 & 0.398 & 0.438 & 0.445 & 0.640 & 0.531\\
\cmidrule{2-12}
&  ATC-NE  & \textemdash & 0.689 & \textbf{0.768} & 0.786 & 0.728 & 0.752 & 0.761 & 0.744 & 0.743 & 0.766\\
&  ATC-MC  & \textemdash & 0.684 & 0.761 & \textbf{0.794} & 0.730 & \textbf{0.761} & \textbf{0.789} & 0.765 & 0.744 & 0.785\\
\bottomrule
\end{tabular}
}
\caption{Full macro $\tau$ metrics for all pairs of training and test domains for BERT models trained on MNLI.}
\label{tab:app_full_macrot_bert_nli}
\end{table*}


\begin{table*}
\resizebox{1\columnwidth}{!}{%
\begin{tabular}{llcccccccccc}
\toprule
& & \multicolumn{10}{c}{Test Domain}\\
\cmidrule{3-12}
Model & Measure & \texttt{books} & \texttt{clothing} & \texttt{home} & \texttt{kindle} & \texttt{movies} & \texttt{pets} & \texttt{sports} & \texttt{tech} & \texttt{tools} & \texttt{toys} \\
\midrule
\multirow{6}{*}{CNN} 
&  NI-SSMBA  & \textbf{0.677} & 0.688 & \textbf{0.758} & \textbf{0.641} & \textbf{0.691} & \textbf{0.786} & \textbf{0.784} & \textbf{0.758} & 0.722 & 0.683\\
&  NI-EDA  & 0.637 & 0.672 & 0.704 & 0.612 & 0.658 & 0.746 & 0.739 & 0.680 & 0.703 & 0.711\\
&  NI-BT  & 0.601 & 0.532 & 0.572 & 0.509 & 0.521 & 0.509 & 0.587 & 0.536 & 0.571 & 0.529\\
&  NI-RandRep  & 0.485 & 0.542 & 0.501 & 0.522 & 0.440 & 0.491 & 0.650 & 0.477 & 0.507 & 0.550\\
\cmidrule{2-12}
&  ATC-NE  & 0.406 & \textbf{0.727} & 0.728 & 0.362 & 0.499 & 0.722 & 0.726 & 0.679 & \textbf{0.730} & \textbf{0.752}\\
&  ATC-MC  & 0.410 & 0.726 & 0.729 & 0.359 & 0.500 & 0.724 & 0.727 & 0.681 & \textbf{0.730} & \textbf{0.752}\\
\midrule
\multirow{6}{*}{BERT} 
&  NI-SSMBA  & 0.790 & \textbf{0.850} & 0.868 & 0.664 & \textbf{0.899} & \textbf{0.864} & \textbf{0.855} & \textbf{0.874} & \textbf{0.880} & 0.845\\
&  NI-EDA  & 0.739 & 0.848 & 0.825 & \textbf{0.731} & 0.793 & 0.836 & 0.825 & 0.828 & 0.837 & 0.841\\
&  NI-BT  & 0.702 & 0.845 & 0.830 & 0.668 & 0.792 & 0.839 & 0.843 & 0.813 & 0.819 & \textbf{0.858}\\
&  NI-RandRep  & \textbf{0.839} & 0.841 & \textbf{0.891} & 0.694 & 0.864 & 0.834 & 0.747 & 0.825 & 0.850 & 0.831\\
\cmidrule{2-12}
&  ATC-NE  & 0.613 & 0.721 & 0.712 & 0.618 & 0.695 & 0.666 & 0.733 & 0.693 & 0.708 & 0.755\\
&  ATC-MC  & 0.621 & 0.735 & 0.730 & 0.616 & 0.711 & 0.691 & 0.749 & 0.710 & 0.726 & 0.764\\
\bottomrule
\end{tabular}
}
\caption{Full micro $\tau$ metrics for all test domains for CNN and BERT models trained on AWS.}
\label{tab:app_full_microt_aws}
\end{table*}

\begin{table*}
\resizebox{1\columnwidth}{!}{%
\begin{tabular}{llcccccccccc}
\toprule
& & \multicolumn{10}{c}{Test Domain}\\
\cmidrule{3-12}
Model & Measure & \texttt{slate} & \texttt{verbatim} & \texttt{facetoface} & \texttt{oup} & \texttt{nineeleven} & \texttt{fiction} & \texttt{telephone} & \texttt{travel} & \texttt{letters} & \texttt{government} \\
\midrule
\multirow{6}{*}{CNN} 
&  NI-SSMBA  & 0.540 & 0.493 & 0.562 & 0.493 & 0.527 & 0.431 & 0.607 & 0.544 & 0.579 & 0.563\\
&  NI-EDA  & 0.560 & 0.555 & 0.427 & 0.520 & 0.488 & 0.435 & 0.499 & 0.552 & 0.532 & 0.542\\
&  NI-BT  & \textbf{0.698} & \textbf{0.684} & \textbf{0.611} & \textbf{0.713} & \textbf{0.713} & \textbf{0.595} & \textbf{0.676} & \textbf{0.714} & \textbf{0.697} & \textbf{0.722}\\
&  NI-RandRep  & 0.448 & 0.417 & 0.427 & 0.503 & 0.472 & 0.324 & 0.395 & 0.450 & 0.550 & 0.509\\
\cmidrule{2-12}
&  ATC-NE  & 0.461 & 0.413 & 0.549 & 0.427 & 0.350 & 0.517 & 0.531 & 0.307 & 0.497 & 0.404\\
&  ATC-MC  & 0.460 & 0.411 & 0.550 & 0.429 & 0.351 & 0.517 & 0.532 & 0.307 & 0.496 & 0.404\\
\midrule
\multirow{6}{*}{BERT} 
&  NI-SSMBA  & \textbf{0.782} & \textbf{0.710} & \textbf{0.742} & 0.652 & \textbf{0.724} & \textbf{0.756} & \textbf{0.779} & 0.691 & \textbf{0.768} & 0.700\\
&  NI-EDA  & 0.563 & 0.556 & 0.481 & 0.622 & 0.586 & 0.545 & 0.529 & 0.607 & 0.607 & 0.630\\
&  NI-BT  & 0.698 & 0.684 & 0.611 & \textbf{0.713} & 0.713 & 0.595 & 0.676 & \textbf{0.714} & 0.697 & \textbf{0.722}\\
&  NI-RandRep  & 0.448 & 0.417 & 0.427 & 0.503 & 0.472 & 0.324 & 0.395 & 0.450 & 0.550 & 0.509\\
\cmidrule{2-12}
&  ATC-NE  & 0.375 & 0.389 & 0.660 & 0.606 & 0.593 & 0.414 & 0.523 & 0.511 & 0.655 & 0.687\\
&  ATC-MC  & 0.405 & 0.434 & 0.676 & 0.632 & 0.614 & 0.466 & 0.554 & 0.530 & 0.663 & 0.693\\
\bottomrule
\end{tabular}
}
\caption{Full micro $\tau$ metrics for all test domains for CNN and BERT models trained on MNLI.}
\label{tab:app_full_microt_nli}
\end{table*}

\begin{table*}
\centering
\scriptsize
\begin{tabular}{lccccc}
\toprule
& \multicolumn{5}{c}{Train Domain (Model)}\\
\cmidrule{2-6}
Measure & SVHN (NiN) & CIFAR10 (NiN) & CIFAR10 (ResNet) & CIFAR10 (VGG) & CINIC10 (CNN)\\
\midrule
NI-RandAug & 0.733 & \textbf{0.797} & 0.746 & \textbf{0.869} & \textbf{0.759}\\
NI-Translate  & \textbf{0.881}  & 0.782 & \textbf{0.833} & 0.699 & 0.329 \\
NI-Erase & 0.324 & 0.409 & 0.708 & -0.183 & 0.606\\
NI-FC & -0.033 & -0.015 & 0.568 & 0.628 & 0.289\\
\midrule
ATC-NE  & 0.859 & 0.648 & 0.757 & 0.773 & 0.548\\
ATC-MC & 0.843 & 0.605 & 0.756 & 0.767 & 0.553\\
\bottomrule
\end{tabular}
\caption{Full in-domain $\tau$ metrics for all test domains for image classification models and datasets.}
\label{tab:app_full_idt_image}
\end{table*}

\begin{table*}
\resizebox{1\columnwidth}{!}{%
\begin{tabular}{llcccccccccc}
\toprule
& & \multicolumn{10}{c}{Train Domain}\\
\cmidrule{3-12}
Model & Measure & \texttt{books} & \texttt{clothing} & \texttt{home} & \texttt{kindle} & \texttt{movies} & \texttt{pets} & \texttt{sports} & \texttt{tech} & \texttt{tools} & \texttt{toys} \\
\midrule
\multirow{6}{*}{CNN} 
&  NI-SSMBA  & 0.646 & \textbf{0.613} & \textbf{0.643} & 0.577 & 0.597 & 0.591 & 0.703 & 0.648 & 0.643 & 0.620\\
&  NI-EDA  & \textbf{0.653} & 0.558 & 0.615 & \textbf{0.660} & 0.447 & \textbf{0.666} & \textbf{0.708} & 0.646 & \textbf{0.663} & 0.565\\
&  NI-BT  & 0.569 & 0.449 & 0.582 & 0.537 & \textbf{0.626} & 0.605 & 0.568 & \textbf{0.653} & 0.640 & \textbf{0.652}\\
&  NI-RandRep  & 0.560 & 0.557 & 0.559 & 0.559 & 0.593 & 0.596 & 0.572 & 0.545 & 0.566 & 0.615\\
\cmidrule{2-12}
&  ATC-NE  & 0.575 & 0.413 & 0.561 & 0.478 & 0.294 & 0.464 & 0.539 & 0.561 & 0.446 & 0.341\\
&  ATC-MC  & 0.566 & 0.405 & 0.558 & 0.478 & 0.300 & 0.461 & 0.540 & 0.569 & 0.451 & 0.338\\
\midrule
\multirow{6}{*}{BERT} 
&  NI-SSMBA  & 0.735 & \textbf{0.874} & 0.868 & 0.693 & 0.780 & 0.871 & \textbf{0.884} & 0.851 & 0.851 & \textbf{0.884}\\
&  NI-EDA  & \textbf{0.875} & 0.788 & 0.814 & \textbf{0.700} & \textbf{0.870} & \textbf{0.894} & 0.818 & \textbf{0.858} & \textbf{0.882} & 0.810\\
&  NI-BT  & 0.801 & 0.811 & 0.857 & 0.638 & 0.835 & 0.813 & 0.860 & 0.846 & 0.803 & 0.750\\
&  NI-RandRep  & 0.839 & 0.841 & \textbf{0.891} & 0.694 & 0.864 & 0.834 & 0.747 & 0.825 & 0.850 & 0.831\\
\cmidrule{2-12}
&  ATC-NE  & 0.733 & 0.724 & 0.822 & 0.640 & 0.721 & 0.711 & 0.778 & 0.783 & 0.799 & 0.783\\
&  ATC-MC  & 0.742 & 0.746 & 0.813 & 0.630 & 0.730 & 0.725 & 0.814 & 0.780 & 0.774 & 0.739\\
\bottomrule
\end{tabular}
}
\caption{Full in-domain $\tau$ metrics for all test domains for CNN and BERT models trained on AWS.}
\label{tab:app_full_idt_aws}
\end{table*}

\begin{table*}
\scriptsize
\centering
\begin{tabular}{llccccc}
\toprule
& & \multicolumn{5}{c}{Train Domain}\\
\cmidrule{3-7}
Model & Measure & \texttt{slate} & \texttt{fiction} & \texttt{telephone} & \texttt{travel} & \texttt{government} \\
\midrule
\multirow{6}{*}{CNN} 
&  NI-SSMBA  & 0.664 & 0.655 & 0.753 & 0.692 & 0.758\\
&  NI-EDA  & 0.629 & 0.723 & \textbf{0.786} & 0.652 & 0.755\\
&  NI-BT  & \textbf{0.751} & \textbf{0.765} & 0.637 & \textbf{0.786} & \textbf{0.826}\\
&  NI-RandRep  & 0.437 & 0.471 & 0.546 & 0.570 & 0.491\\
\cmidrule{2-7}
&  ATC-NE  & 0.704 & 0.662 & 0.719 & 0.715 & 0.722\\
&  ATC-MC  & 0.703 & 0.660 & 0.722 & 0.716 & 0.728\\
\midrule
\multirow{6}{*}{BERT} 
&  NI-SSMBA  & 0.713 & 0.754 & 0.766 & 0.785 & \textbf{0.839}\\
&  NI-EDA  & 0.608 & 0.664 & 0.781 & 0.575 & 0.726\\
&  NI-BT  & \textbf{0.751} & 0.765 & 0.637 & \textbf{0.786} & 0.826\\
&  NI-RandRep  & 0.437 & 0.471 & 0.546 & 0.570 & 0.491\\
\cmidrule{2-7}
&  ATC-NE  & 0.722 & \textbf{0.796} & 0.740 & 0.737 & 0.701\\
&  ATC-MC  & 0.745 & 0.791 & \textbf{0.791} & 0.719 & 0.696\\
\bottomrule
\end{tabular}
\caption{Full in-domain $\tau$ metrics for all train domains for CNN and BERT models trained on MNLI.}
\label{tab:app_full_idt_nli}
\end{table*}

\begin{table}[h]
\begin{subtable}{1\textwidth}
\scriptsize
\centering
\scriptsize
\resizebox{0.45\columnwidth}{!}{%
\setlength\tabcolsep{4pt}
\begin{tabular}{lccccc}
\toprule
& \multicolumn{2}{c}{Domain Shifts} & \multicolumn{2}{c}{ImageNet-A}\\
 \cmidrule(lr){2-3}
 \cmidrule(lr){4-5}
Measure & $R^2$ & Macro $\tau$ &  $R^2$ & Macro $\tau$ & ID $\tau$\\
\midrule
NI-RandAug & 0.091 & 0.061 & --- & --- & --- \\
NI-Translate & 0.159 & 0.099 & --- & --- & --- \\
NI-Erase & 0.174 & 0.153 & --- & --- & ---  \\
NI-FC & 0.139 & 0.095 & --- & --- & ---  \\
\midrule
ATC-NE & 0.272 & 0.136 & --- & --- & --- \\
ATC-MC & 0.279 & 0.135 & --- & --- & --- \\
\bottomrule
\end{tabular}
}
\caption{Standard deviations for ImageNet scale results. No standard deviations are reported for ImageNet-A or ID $\tau$ since we report only a single value.}
\label{tab:appimagenet_corr}
\vspace{0.3cm}
\end{subtable}


\begin{subtable}{1\textwidth}
\centering
\scriptsize
\resizebox{0.65\columnwidth}{!}{%
\begin{tabular}{lcccccc}
\toprule
& \multicolumn{3}{c}{CI10} & \multicolumn{3}{c}{Numbers}\\
 \cmidrule(lr){2-4}
 \cmidrule(lr){5-7}
Measure & $R^2$ & Macro $\tau$ & ID $\tau$ & $R^2$ & Macro $\tau$ & ID $\tau$ \\
\midrule
NI-RandAug & 0.044 & 0.054 & 0.048 & 0.021 & 0.026 & ---\\
NI-Translate & 0.170 & 0.181 & 0.197 & 0.043 & 0.032 & ---\\
NI-Erase & 0.254 & 0.348 & 0.345 & 0.065 & 0.033 & ---\\
NI-FC & 0.239 & 0.225 & 0.255 & 0.074 & 0.083 & ---\\
\midrule
ATC-NE & 0.169 & 0.115 & 0.091 & 0.109 & 0.059 & ---\\
ATC-MC & 0.168 & 0.101 & 0.093 & 0.077 & 0.035 & ---\\
\bottomrule
\end{tabular}
}
\caption{Standard deviations for small scale image results. No standard deviation is reported for Numbers ID $\tau$ since we report only a single value.}
\label{tab:app_ic_corr}
\vspace{0.5cm}
\end{subtable}

\begin{subtable}{1\textwidth}
\scriptsize
\centering
\resizebox{0.8\columnwidth}{!}{%
\setlength\tabcolsep{4pt}
\begin{tabular}{lcccccccc}
\toprule
& \multicolumn{4}{c}{CNN} & \multicolumn{4}{c}{RoBERTa}\\
 \cmidrule(lr){2-5}
 \cmidrule(lr){6-9}
Measure & $R^2$ & Macro $\tau$ & Micro $\tau$ & ID $\tau$ &  $R^2$ & Macro $\tau$ & Micro $\tau$ & ID $\tau$\\
\midrule
NI-SSMBA & 0.079 & 0.067 & 0.046 & 0.035 & 0.012 & 0.063 & 0.064 & 0.065 \\
NI-EDA & 0.108 & 0.051 & 0.041 & 0.072 & 0.016 & 0.058 & 0.040 & 0.055\\
NI-BT & 0.161 & 0.078 & 0.032 & 0.060 & 0.021 & 0.060 & 0.061 & 0.063\\
NI-RandRep & 0.069 & 0.080 & 0.054 & 0.021 & 0.015 & 0.060 & 0.055 & 0.055 \\
\midrule
ATC-NE & 0.126 & 0.110 & 0.143 & 0.092 & 0.168 & 0.135 & 0.044 & 0.051 \\
ATC-MC & 0.125 & 0.109 & 0.143 & 0.091 & 0.139 & 0.112 & 0.048 & 0.050  \\
\bottomrule
\end{tabular}
}
\caption{Standard deviations on sentiment analysis results.}
\label{tab:app_sa_corr}
\vspace{0.5cm}
\end{subtable}
\begin{subtable}{1\textwidth}
\scriptsize
\centering
\resizebox{0.8\columnwidth}{!}{%
\setlength\tabcolsep{4pt}
\begin{tabular}{lcccccccc}
\toprule
& \multicolumn{4}{c}{CNN} & \multicolumn{4}{c}{RoBERTa}\\
 \cmidrule(lr){2-5}
 \cmidrule(lr){6-9}
Measure & $R^2$ & Macro $\tau$ & Micro $\tau$ & ID $\tau$ &  $R^2$ & Macro $\tau$ & Micro $\tau$ & ID $\tau$ \\
\midrule
NI-SSMBA & 0.098 & 0.060 & 0.048 & 0.043 & 0.035 & 0.046 & 0.040 & 0.041\\
NI-EDA & 0.107 & 0.068 & 0.046 & 0.060 & 0.108 & 0.087 & 0.044 & 0.075 \\
NI-BT & 0.045 & 0.049 & 0.042 & 0.063 & 0.045 & 0.049 & 0.042 & 0.063  \\
NI-RandRep & 0.196 & 0.079 & 0.061 & 0.049 & 0.196 & 0.079 & 0.061 & 0.049 \\
\midrule
ATC-NE & 0.108 & 0.068 & 0.076 & 0.022 & 0.206 & 0.180 & 0.111 & 0.032 \\
ATC-MC & 0.108 & 0.068 & 0.076  & 0.024 & 0.174 & 0.164 & 0.100 & 0.038 \\
\bottomrule
\end{tabular}
}
\caption{Standard deviations on NLI results.}
\label{tab:app_nli_corr}
\vspace{0.3cm}
\end{subtable}

\caption{Standard deviations for reported values in Table \ref{tab:res_full}}
\label{tab:app_res_full_std}
\end{table}

\end{document}